\documentclass{article}
\usepackage[colorlinks=true, allcolors=blue]{hyperref}
\usepackage{apacite}

\usepackage{authblk}

\usepackage[english]{babel}

\usepackage[letterpaper,top=2cm,bottom=2cm,left=3cm,right=3cm,marginparwidth=1.75cm]{geometry}

\usepackage{amsmath}
\usepackage{graphicx}

\title{The Current State of Summarization}
\author{Fabian Retkowski}
\affil{Karlsruhe Institute of Technology (KIT) \\ \texttt{\href{mailto:retkowski@kit.edu}{retkowski@kit.edu}}} 

\begin{document}
\maketitle

\begin{abstract}
\noindent
With the explosive growth of textual information, summarization systems have become increasingly important. This work aims to concisely indicate the current state of the art in abstractive text summarization. As part of this, we outline the current paradigm shifts towards pre-trained encoder-decoder models and large autoregressive language models. Additionally, we delve further into the challenges of evaluating summarization systems and the potential of instruction-tuned models for zero-shot summarization. Finally, we provide a brief overview of how summarization systems are currently being integrated into commercial applications.
\end{abstract}

\section{Introduction}

Summarization is the process of extracting the most important information from a text and presenting it in a condensed form. With vast amounts of information produced at an unprecedented rate, organizations and individuals alike face unique challenges, heightening the demand for effective summarization systems. For researchers of many fields, it is challenging to keep up with the latest developments in their field including Artificial Intelligence itself as vicariously indicated by the number of journal publications per year which has almost tripled since 2015 \shortcite{zhang_ai_2022}.

In general, two different forms of summarization are distinguished: extractive and abstractive. In extractive summarization, the system is tasked with selecting passages from the document to be included in the summary. Abstractive summarization, on the other hand, aims to rephrase the most important aspects of a document with a different syntax. As language models are becoming more and more capable, research is increasingly shifting from extractive to abstractive summarization, which is considered more challenging, but also more fluent, diverse, and readable.

\begin{figure}
    \centering
    \includegraphics[trim={6cm 2.25cm 6cm 3cm},clip,scale=0.45]{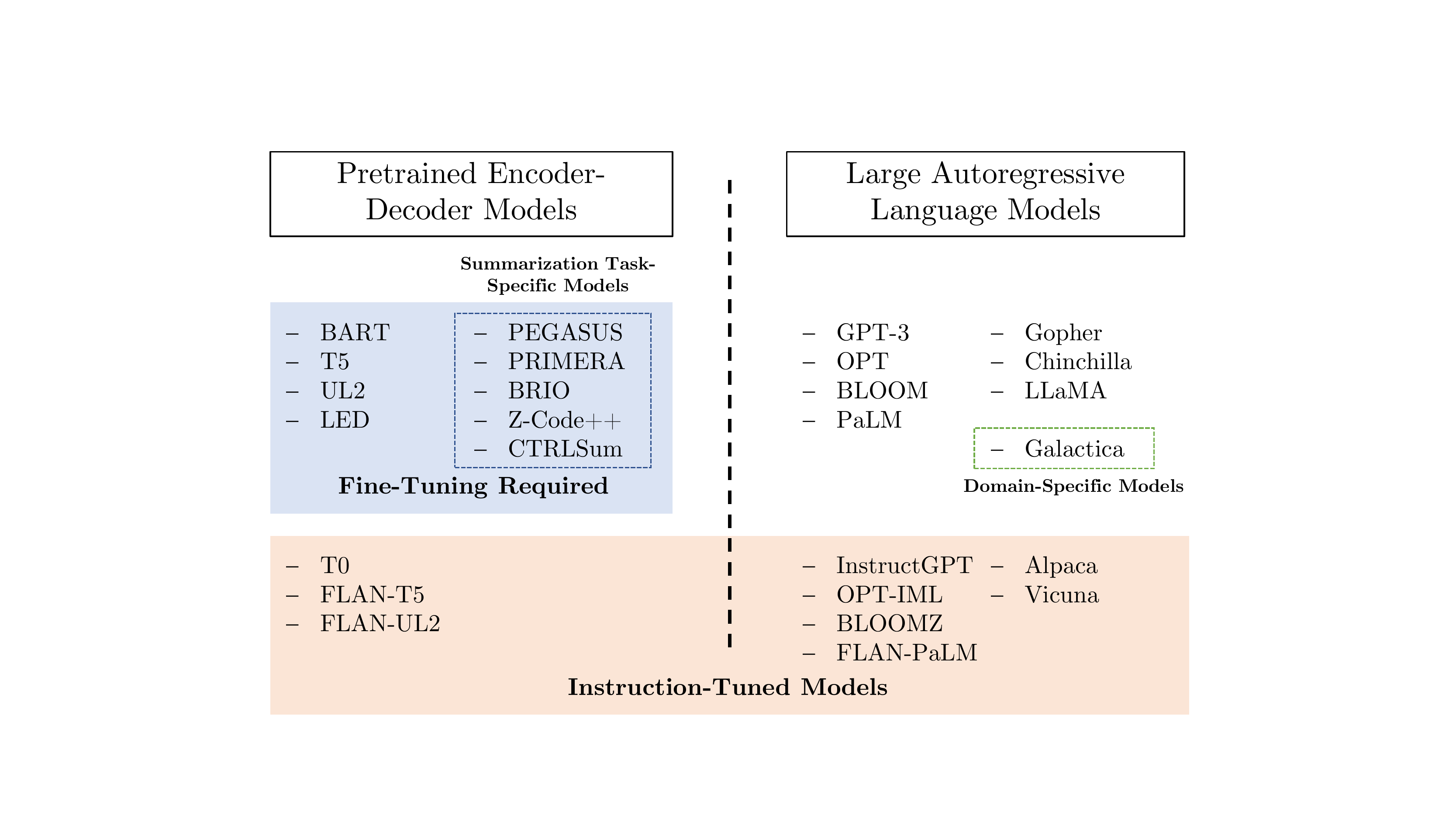}
    \caption{Current summarization systems can be broadly divided into pre-trained encoder-decoder models and large autoregressive language models. In general, instruction-tuned models are most capable when it comes to zero-shot summarization. Other encoder-decoder models usually require fine-tuning, while autoregressive LLMs are less effective without instruction tuning. Some pre-trained encoder-decoder models are specifically designed for the summarization task.}
    \label{fig:languagemodels}
\end{figure}

This paper covers recent advances in abstractive text summarization, with a focus on pre-trained encoder-decoder models (Section \ref{section:enc-dec}), large autoregressive language models (Section \ref{section:llm}), and instruction-tuned variants (Section \ref{section:instruct}). While aiming to be reasonably comprehensive, Figure \ref{fig:languagemodels} gives an overview of the covered models. In Section \ref{sec:eval}, current evaluation protocols are discussed in the context of the paradigm shift towards large language models. At the end of the paper, we discuss limitations, potentials (Section \ref{sec:limit}), and current commercialization efforts (Section \ref{sec:commerz}).

%  trim={<left> <lower> <right> <upper>}

\section{Pre-Trained Encoder-Decoder Models}
\label{section:enc-dec}

\begin{figure}
    \centering
    \includegraphics[width=\textwidth]{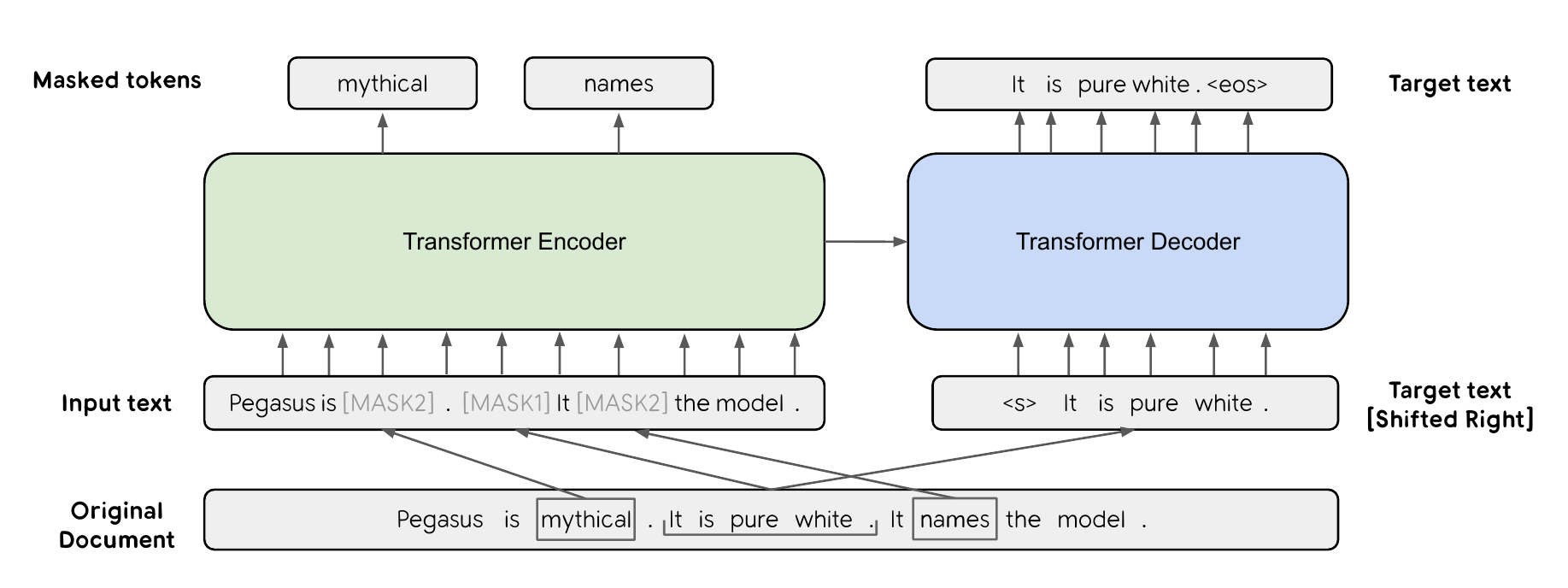}
    \caption{The PEGASUS architecture with its pre-training objectives. The model combines Masked Language Modeling (MLM) as well as Gap Sentences Generation (GSG). As part of GSG, important sentences are masked and used as a target for the decoder. The importance is proximately scored by ROUGE-1 between a sentence and the remaining portions of the document. \protect\shortcite{zhang_pegasus_2020}}
    \label{fig:pegasus}
\end{figure}

Pre-trained encoder-decoder models have gained tremendous popularity in recent years and are now widely established in the field of natural language processing. These models are trained in a self-supervised setting on a large, unlabeled corpus. Notable examples include models such as the denoising autoencoder BART \shortcite{lewis_bart_2020} and T5 \shortcite{raffel_exploring_2020} that is trained on a fill-in-the-blank objective. UL2 \shortcite{tay_ul2_2022} serves as a more recent example that generalizes and combines several denoising pre-training objectives. By fine-tuning these models on task-specific datasets, they have achieved state-of-the-art results across many tasks including summarization. Some pre-trained models are specifically designed for the task of summarization by choosing a pre-training objective that resembles summarization. For example, in Figure \ref{fig:pegasus}, the architecture of PEGASUS \shortcite{zhang_pegasus_2020} can be observed, which is trained by removing important sentences from the input document and tasking the model with regenerating them. In a comprehensive evaluation of 23 models for the summarization task, \shortciteA{fabbri_summeval_2021} conclude that \mbox{PEGASUS}, BART, and T5 "consistently performed the best on most dimensions", which involves human evaluations as well as automatic metrics. Recently, a task-specific fine-tuning mechanism called BRIO \shortcite{liu_brio_2022} was proposed for summarization. This method introduces a contrastive learning component to prevent assigning the entire distribution mass to the reference summary and instead account for candidate summaries as well. BRIO has been applied to several models, including BART and \mbox{PEGASUS}. Another noteworthy model is \mbox{Z-Code++} \shortcite{he_z-code_2023}, as it incorporates an intermediate task-adaptive fine-tuning step using a broad collection of summarization datasets before fine-tuning on a specific summarization task. This method has been shown to be especially effective in low-resource settings.

\section{Large Autoregressive Language Models}
\label{section:llm}

Another significant paradigm shift is the recent emergence of large autoregressive language models (LLMs). These decoder-only models tend to have many more parameters and are trained using the traditional causal language modeling objective of predicting the next token in a sequence. \shortciteA{brown_language_2020} were the first to demonstrate that this approach, at scale, enables zero-shot prompting to perform a wide variety of downstream tasks. Without any gradient updates, this involves priming the model with a task-specific natural language prompt (e.g., "Question: \textlangle question\textrangle\ Answer:") and then producing an output by sampling from the model. The same paradigm also allows for zero-shot summarization, which can be achieved by appending "TL;DR:" to a prompt, among other options.

The most popular model in this category is GPT-3 \shortcite{brown_language_2020} with its 175B parameters. OPT \shortcite{zhang_opt_2022} and BLOOM \shortcite{bigscience_workshop_bloom_2022} are two open-source alternatives aimed to replicate the results. Gopher \shortcite{rae_scaling_2022} and PaLM \shortcite{chowdhery_palm_2022} take this approach to the extreme by scaling to even larger model sizes of up to 560B parameters. On the contrary, Chinchilla \shortcite{hoffmann_training_2022} and LLaMA \shortcite{touvron_llama_2023} take scaling laws and compute budgets more strictly into consideration and this way achieve training a 70B respectively 65B model while still being able to match or outperform larger models. It is also worth mentioning the Galactica 120B scientific language model \shortcite{taylor_galactica_2022}, which demonstrates the effectiveness of specialized LLMs. It outperforms other LLMs in its specific domain by using a sophisticated dataset design that incorporates domain-adapted tokenization. It treats citations and modalities, like chemical formulas and protein sequences, specially by introducing task-specific tokens for them.

\section{Instruction-Tuned Models}
\label{section:instruct}

\begin{figure}
    \centering
    \includegraphics[trim={0 9.5cm 25cm 3cm},clip,scale=0.3]{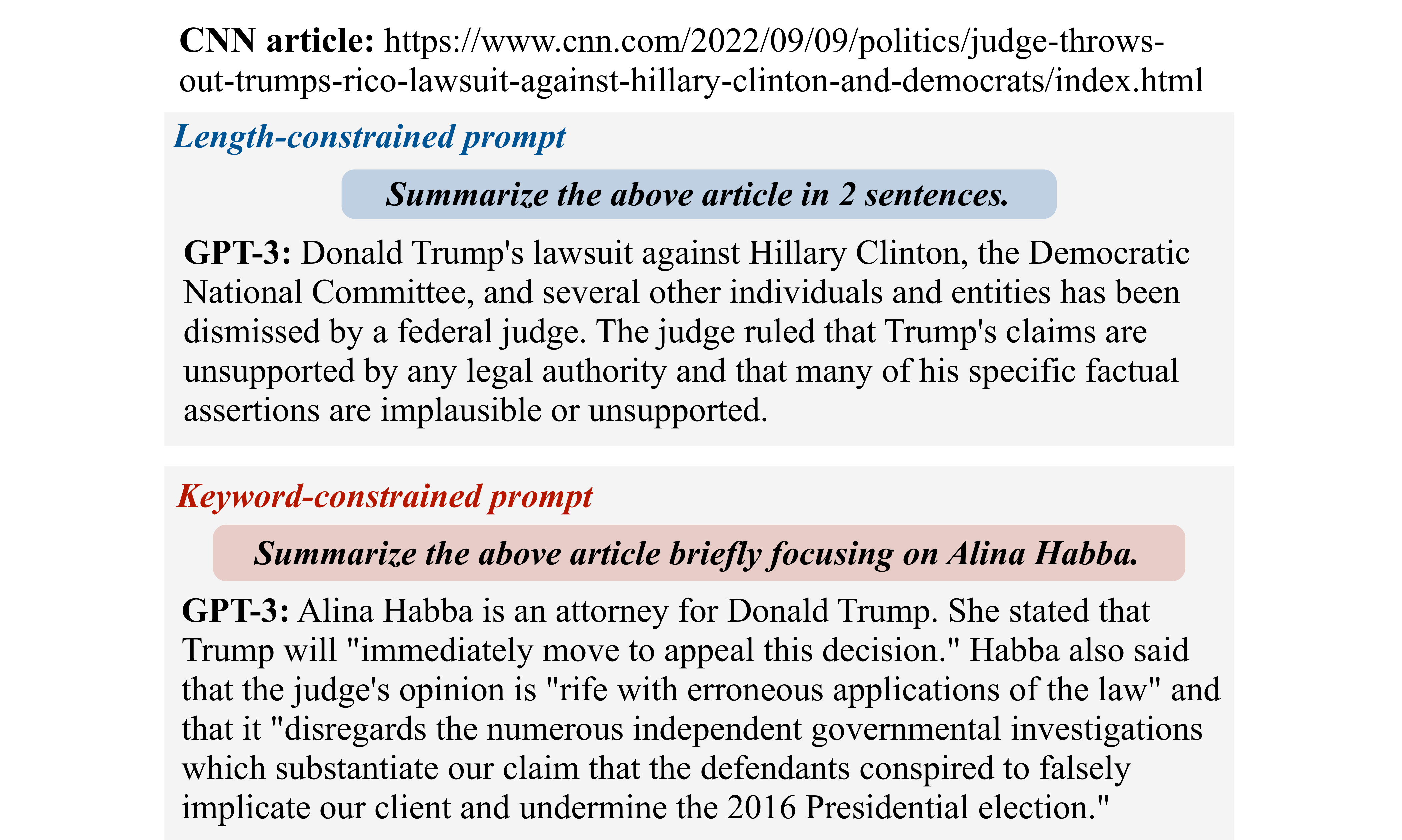}
    \caption{Exemplary instructions for zero-shot summarization using GPT-3. Notably, the natural language instructions of LLMs enable greater control over tasks, such as length-constrained summarization. \protect\shortcite{goyal_news_2022}}
    \label{fig:prompt}
\end{figure}

Instruction tuning refers to the process of fine-tuning a pre-trained model with a diverse range of datasets that are described using natural language task instructions. This step ensures that the training process is more aligned with how the model will be used during inference and has been shown to significantly improve performance on zero-shot tasks. It enables the model to be straightforwardly and more reliably instructed to perform a certain task. For instance, it is now possible to use "Summarize the article: \textlangle article\textrangle" as a prompt for the summarization task. More prompt examples are shown in Figure \ref{fig:prompt}. To tune models for instructions, the most common approaches are supervised fine-tuning and reinforcement learning from human feedback \shortcite<RLHF, >{christiano_deep_2017}. When it comes to pre-trained encoder-decoder models, there are several popular instruction-tuned models available. For instance, T0 \shortcite{sanh_multitask_2022} and FLAN-T5 \shortcite{chung_scaling_2022}, which are both based on T5, have gained significant traction among practitioners. The same is true for large autoregressive language models of which most have an instruction-tuned equivalent: InstructGPT \shortcite{ouyang_training_2022}, OPT-IML \shortcite{iyer_opt-iml_2023}, BLOOMZ \shortcite{muennighoff_crosslingual_2023}, FLAN-PaLM \shortcite{chung_scaling_2022}. \shortciteA{taylor_galactica_2022} demonstrated with Galactica an alternative approach to enable rudimentary instruction prompting with their prompt pre-training method. This involves adding task prompts to the pre-training, rather than tuning the model after pre-training. A recent trend in the open-source community is to fine-tune LLMs based on conversational and instruction-following data generated by an existing and strong instruction-tuned LLM such as ChatGPT. This has led to the development of Alpaca and Vicuna, both of which are based on LLaMA \shortcite{taori_alpaca_2023,the_vicuna_team_vicuna_2023,wang_self-instruct_2023}. The task of summarization is represented in most natural-language-prompted datasets. For example, in the API prompt dataset used by InstructGPT, 4.2\% of instructions fall under the 'summarization' use case. Similarly, T0 augments classic summarization datasets like CNN Daily Mail \shortcite{nallapati_abstractive_2016} or SamSum \shortcite{gliwa_samsum_2019} with instruction templates that can be used to fine-tune the model.

\section{Evaluation of Large Language Models}
\label{sec:eval}

Most commonly, summarization systems are evaluated on automated metrics. ROUGE \cite{lin_rouge_2004} in particular has a long-standing history in the field and measures the lexical overlap between reference summaries and generated summaries. More recent metrics such as BertScore \shortcite{zhang_bertscore_2019} and BARTScore \shortcite{yuan_bartscore_2021}, which are better at capturing semantic equivalence, are also becoming increasingly established. However, as large language models become more capable and generalize to a wide range of tasks, they are less frequently or thoroughly evaluated on summarization tasks specifically. Instead, they are evaluated on benchmark suits that focus on question answering and common-sense reasoning, such as SuperGLUE \shortcite{wang_superglue_2019} or MMLU \shortcite{hendrycks_measuring_2020}, that do not explicitly involve summarization. As a result, several research groups have independently investigated the capabilities and limitations of LLMs in summarization more recently \shortcite{goyal_news_2022,bhaskar_prompted_2023,liu_revisiting_2023,qin_is_2023,xiao_chatgpt-steered_2023,yang_exploring_2023,zhang_benchmarking_2023}. According to \shortciteauthor{goyal_news_2022}, summaries generated by instruction-tuned GPT-3 receive lower scores on automatic metrics compared to fine-tuned encoder-decoder models (T0 and BRIO). Despite this, the model outperforms them significantly in human evaluation. The conducted human evaluation by \shortciteauthor{zhang_benchmarking_2023} suggests that they even surpass the reference summaries in quality and are on par with high-quality summaries collected separately for this evaluation. These works cast great doubt on existing evaluation protocols, especially in the context of this paradigm shift. Several of the works describe the low correlation of automatic metrics with human judgment, low reference quality, lacking inter-annotator agreement, and different summarization styles (in length, abstractiveness, formality) as problematic. This is in line with issues raised in previous works such as \shortciteA{fabbri_summeval_2021} that point out the lack of comparability of summarization evaluation protocols -- for automated metrics and human evaluation alike. Considering these issues and with summarization systems rivaling human performance, \shortciteauthor{zhang_benchmarking_2023} hypothesize that a limit is reached in evaluating "single-document news summarization", while \shortciteauthor{yang_exploring_2023} call for "rethinking further directions for various text summarization tasks". In fact, the "glass ceiling" phenomenon has been observed more broadly in natural language generation, with even recent automated metrics barely improving correlation with human judgment \shortcite{colombo_glass_2022}.

\section{Limitations and New Frontiers}
\label{sec:limit}

As discussed, there are severe limitations to the current evaluation metrics and protocols, and finding a new standard is an essential area for future research. \shortciteA{liu_revisiting_2023}, for example, suggest using atomic facts to reduce ambiguity in human evaluation, while a recent work in the area of machine translation shows that LLMs themselves make state-of-the-art evaluators offering greater correlation with human judgment than any other automatic metric \shortcite{kocmi_large_2023}. The latter is also supported by \shortciteA{kadavath_language_2022}, who find that LLMs are capable of self-evaluation. At the same time, LLMs are known to suffer from hallucinations \shortcite{ji_survey_2023} and as summarization moves to higher levels of abstractiveness, factuality comes into question. Works like \shortciteA{bhaskar_prompted_2023} or \shortciteA{goyal_news_2022} show that summarization factuality is still an unsolved issue for LLMs, while others openly discuss how to measure factuality in the first place \shortcite{kryscinski_evaluating_2020,pagnoni_understanding_2021}.

\paragraph{Long Document Summarization.}

Despite exponential progress (see Figure \ref{fig:context}), many current summarization systems are still hindered by the limited context windows of language models which prevent them from processing longer documents that would especially benefit from summarization such as lengthy news articles, scientific papers, podcasts, or books. There are several common strategies to overcome this limitation. One simple method involves truncating the input text \shortcite{zhao_seal_2020,wang_squality_2022}. For some document types such as news articles, this might serve as a reasonable strategy, as they tend to convey the most salient information in the beginning. In fact, selecting the first $k$ sentences (Lead-$k$) is often used as a baseline summary for news summarization systems \shortcite{see_get_2017,zhong_searching_2019}. In a similar vein, for the summarization of scientific papers, often only the abstract, introduction, and conclusion (AIC) are passed to the summarizer, as previous research found these sections to be the most salient \shortcite{sharma_bigpatent_2019,cachola_tldr_2020}. Another approach is to employ an extractive summarizer or retrieval module such as \shortciteA<Dense Passage Retriever, >{karpukhin_dense_2020}, as part of a two-stage system, to select important segments before passing the text to the abstractive summarizer \shortcite{liu_text_2019,ladhak_exploring_2020,wang_squality_2022}. There are also transformer architectures that do not suffer from these limitations such as LED \shortcite{beltagy_longformer_2020} or LongT5 \shortcite{guo_longt5_2022} which replace $O(n^2)$ attention patterns with more efficient ones. Finally, experiments have been conducted on summarizing chunks of the text in potentially multiple iterations before producing a final, coherent summary \shortcite{gidiotis_divide-and-conquer_2020,zhao_seal_2020,wu_recursively_2021,zhang_summn_2022,yang_exploring_2023}.

\paragraph{Multi-Document Summarization.} The process of creating a summary from a collection of documents related to a specific topic is called multi-document summarization (MDS). This presents similar challenges to summarizing a long document, as the problem of limited context length is amplified when multiple documents are involved. Understanding the relationships between the documents is also essential for completing the task effectively. The first strategy for MDS is to simply concatenate all documents into one large text and use techniques designed for single-document summarization. However, this requires the model to process very long sequences. Therefore, a two-stage process similar to that used for long document summarization is commonly employed \shortcite{liu_generating_2018,liu_hierarchical_2019}. State-of-the-art approaches also use hierarchical architectures or graph-based methods to capture inter-document relations \shortcite{liu_hierarchical_2019,li_leveraging_2020,pasunuru_efficiently_2021}. At the same time, MDS approaches increasingly aim to utilize pre-trained encoder-decoder models such as BART, T5, or PEGASUS \shortcite{goodwin_flight_2020,pasunuru_efficiently_2021}. One recent and noteworthy model in this category, PRIMERA, is specifically designed for MDS and builds upon the foundations laid by PEGASUS \shortcite{xiao_primera_2022}. For the GSG objective, PRIMERA chooses sentences that represent clusters of documents. It employs a document concatenation approach and architecturally uses LED to handle long sequences. In this manner, the model is generally applicable, with no dependencies on specific datasets. Although there is no scientific evaluation yet, the recent emergence and popularity of practical tools like LangChain\footnote{\url{https://github.com/hwchase17/langchain}} and LlamaIndex\footnote{\url{https://github.com/jerryjliu/llama_index}} hint towards the use of LLMs to handle collections of documents. For instance, LlamaIndex enables the storage of documents in an index that is organized like a tree, with each node representing a summary of its child nodes.

\paragraph{Controllable Summarization.}

Controllable summarization is a multifaceted research question that refers to both the form or style (such as length, formality, or abstractiveness) and the content of a summary. The summary may be conditioned on a specific aspect or entity or, more broadly, on any given keyword or query. In recent years, a wide variety of approaches have been proposed. One of the most comprehensive systems is CTRLSum \shortcite{he_ctrlsum_2022}, a pre-trained encoder-decoder that generalizes controllability by utilizing keywords and prompts alike. In evaluations, the authors show the effectiveness of their method for length and entity control, as well as some more specialized tasks (e.g., patent purpose summarization). Recent studies conducted by \shortciteA{goyal_news_2022}, \shortciteA{xiao_chatgpt-steered_2023}, and \shortciteA{yang_exploring_2023} offer initial insights into the potential of instruction-tuned LLMs like GPT-3 and ChatGPT. These systems have shown great promise for diverse summarization tasks based on keywords, aspects, and queries. Figure \ref{fig:prompt} shows two examples of how zero-shot prompting can enable controllable summarization in such systems. Nevertheless, the potential of LLMs for this task is still largely unexplored. \shortciteA{yang_exploring_2023} notes that their results can only serve as a lower bound, as the models are naively prompted without any prompt tuning or self-correction. A first glimpse of the potential of a more sophisticated prompting strategy is provided by \shortciteA{xiao_chatgpt-steered_2023} who suggest editing generated summaries with an editor model based on instructions from a separately trained model. In stark contrast, there is also a significant amount of research that focuses on controlling only one aspect of summarization. For example, in length-controllable summarization alone, systems have been proposed that early-stop the decoding process \shortcite{kikuchi_controlling_2016}, select information before passing it to the summarizer \shortcite<LPAS;>{saito_length-controllable_2020}, or incorporate length information as part of the input \shortcite{kikuchi_controlling_2016,liu_controlling_2018}. More recently, \shortciteA{liu_length_2022} also introduced a length-aware attention mechanism (LAAM).

\paragraph{Multi-Modal Summarization.} So far, most research attention has been given to text summarization systems. However, there is an abundance of media and content such as podcasts, movies, and meetings that not only involve text but also other modalities including images, videos, and audio. These other modalities potentially contain key information that a pure text summarization system might miss, thus creating a semantic gap. For instance, \shortciteA{li_multi-modal_2017} have demonstrated the importance of including audio and video information in the task of summarizing multimedia news, while the work of \shortciteA{li_keep_2019} has shown the value of including participants' head orientation and eye gaze when summarizing meetings. One of the key challenges of multi-modal summarization systems is the fusion of different input modalities. Currently, most systems take a late-fusion approach \shortcite<see>{jangra_survey_2023}, for example by utilizing a pre-trained encoder. However, recently, a number of promising Transformer-based models have been proposed, which allow the input of diverse modalities such as Perceiver IO \shortcite{jaegle_perceiver_2021} or GATO \shortcite{reed_generalist_2022} that have yet to be applied for the summarization task.

\section{Commercialization}
\label{sec:commerz}

\begin{figure}
    \centering
    \includegraphics[clip,scale=0.45]{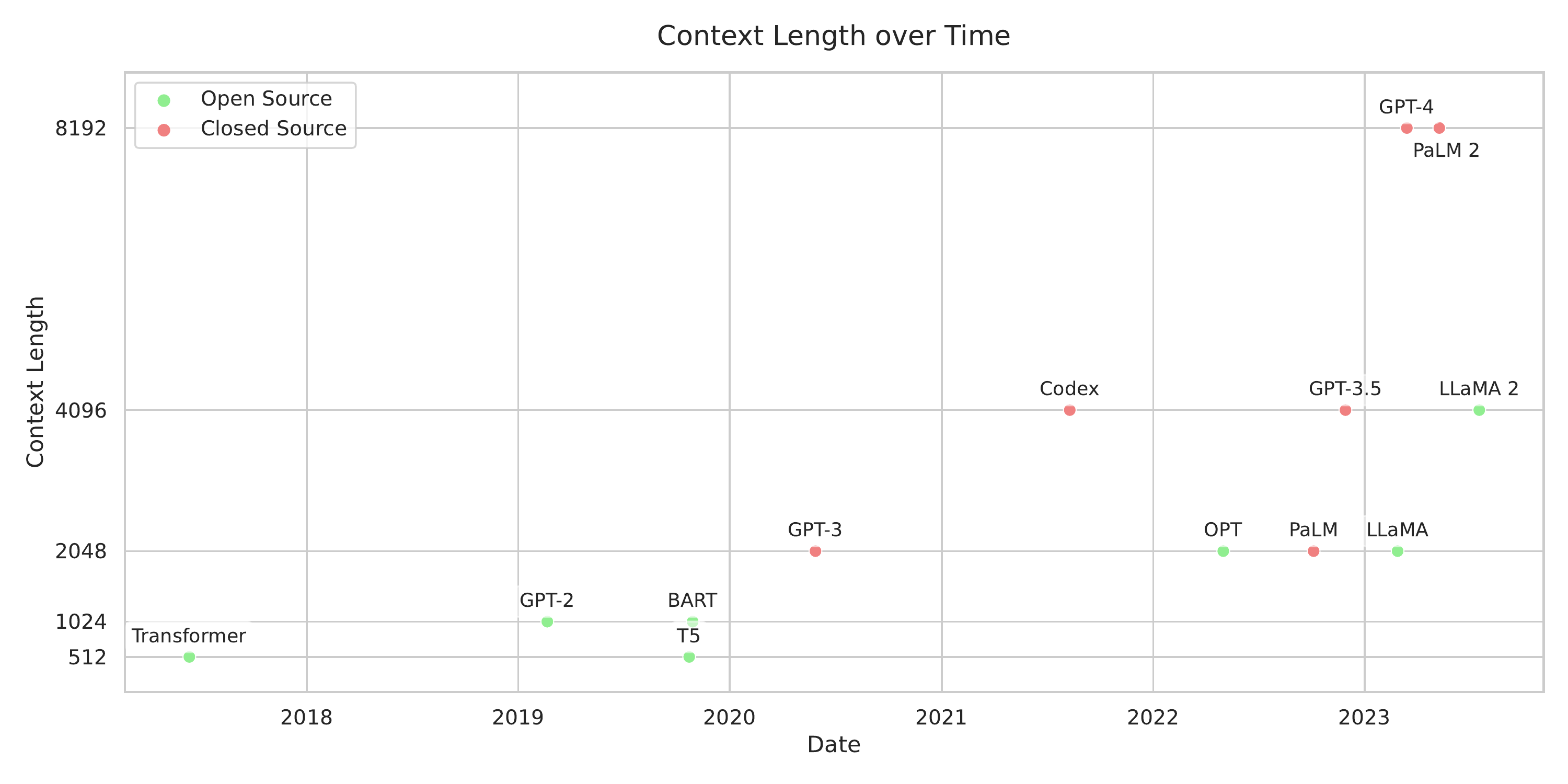}
    \caption{The context length has been steadily and exponentially increasing in open-source and closed-source language models alike. Not considered are models like LED, which specifically try to maximize the context length at the cost of performance otherwise.}
    \label{fig:context}
\end{figure}

With language models having surpassed a certain level of performance, the creation and integration of these models into products and tools have become increasingly common, leading to a "gold rush" of NLP startups
\cite{butcher_heres_2022,toews_wave_2022}. For summarization systems in particular, the context lengths of models are of utmost importance and have expanded exponentially in recent years as can be seen in Figure \ref{fig:context}, to a level that is practical for more tasks and commercially viable. As such, many summarization systems have become productized and have been made available in consumer-oriented interfaces over the past year. In 2022, Google introduced document summarization in Google Docs \shortcite{saleh_auto-generated_2022} and conversation summarization in Google Chat \shortcite{saleh_conversation_2022}, both powered by fine-tuning the PEGASUS model. However, low-quality summaries in the datasets are mentioned as problematic. To tackle this issue, the developers utilize techniques such as dataset distillation, data formatting, and clean-ups, while continuing to collect more training data. Through knowledge distillation, they distill the models into more efficient hybrid architectures of a transformer encoder and a recurrent neural network (RNN) decoder. Separately, an additional model is trained to filter out generated summaries that are of low quality. More recently, Microsoft announced plans to roll out meeting summarization powered by GPT-3.5 in Microsoft Teams in Q2 2023 \shortcite{herskowitz_microsoft_2023}, but they have not provided any further technical details. Discord, the community messaging platform, uses "OpenAI technology" for grouping messages into topics for conversation summaries \cite{midha_discord_2023}. Zoom's recent smart recording feature, which includes meeting summarization and smart chaptering, vaguely mentions the use of GPT-3 to "augment" its own models \cite{parthasarathy_zooms_2023}. Cohere just launched a dedicated text summarization endpoint \shortcite{hillier_introducing_2023} that largely avoids several problems of LLMs such as the need for prompt engineering and limited context length. In addition, they offer settings to gain more control over the generated summaries: the level of extractiveness, the length, and the format (either fluent text or bullet points). More broadly, access to any standard LLM naturally allows for summarization by specifying the respective prompt. This is true for OpenAI's GPT-3, AI21 Studio, Antrophic's Claude, or Cohere Generate -- to name some that are available via paid APIs and power summarization functionalities in many commercial applications. ChatGPT might be notable, as it also enables a more interactive approach to summarization. Domain-specific summarization tools are another area of interest. For instance, Zoom IQ for Sales \cite{larkin_zoom_2022} aims to provide insights and summaries for sales meetings, while BirchAI, a spinoff from the Allen Institute for Artificial Intelligence, focuses solely on providing customer call summaries for call centers. Meanwhile, beyond big tech and distinguished AI labs, summarization systems are starting to reach many more surfaces such as browsers \shortcite<Opera;>{szyndzielorz_opera_2023}, email clients \shortcite<Shortwave;>{wenger_ai_2023} or note-taking apps \shortcite<Notion;>{zhao_notion_2023}. This trend suggests that summarization is not an application on its own, but a basic feature to be widely implemented on most surfaces and to be widely accessible in the foreseeable future.

\section{Conclusion}
Text summarization is a rapidly evolving field with two recent paradigm shifts. First, towards fine-tuning pre-trained encoder-decoder models, and second and even more recently, towards zero-shot prompting of instruction-tuned language models. As a result of these developments, it appears that single-document summarization has reached a tipping point where the focus on improving automated metrics has diminishing returns and might even misdirect the research community. Therefore, we suggest a shift of emphasis towards improving human evaluation protocols and exploring self-evaluation of LLMs. Additionally, more targeted evaluation of certain aspects, such as factuality, should be considered and more broadly the uncovering of capabilities of pre-trained language models and fine-tuned summarization models. However, when contemplating summarization in a wider scope, tasks such as multi-document summarization and multi-modal summarization continue to present significant hurdles. Nonetheless, abstractive text summarization systems for single documents have matured and are rapidly being integrated into consumer products.

\section*{Acknowledgement}
The project on which this report is based was funded by the Volkswagen Stiftung.

\bibliographystyle{apacite}
\bibliography{references}

\begin{thebibliography}{}

\bibitem [\protect \citeauthoryear {%
Beltagy%
, Peters%
\BCBL {}\ \BBA {} Cohan%
}{%
Beltagy%
\ \protect \BOthers {.}}{%
{\protect \APACyear {2020}}%
}]{%
beltagy_longformer_2020}
\APACinsertmetastar {%
beltagy_longformer_2020}%
\begin{APACrefauthors}%
Beltagy, I.%
, Peters, M\BPBI E.%
\BCBL {}\ \BBA {} Cohan, A.%
\end{APACrefauthors}%
\unskip\
\newblock
\APACrefYearMonthDay{2020}{{\APACmonth{12}}}{}.
\newblock
\APACrefbtitle {Longformer: {The} {Long}-{Document} {Transformer}.}
  {Longformer: {The} {Long}-{Document} {Transformer}.}
\newblock
\APACaddressPublisher{}{arXiv}.
\newblock
\begin{APACrefURL} [{2023-03-03}]\url{http://arxiv.org/abs/2004.05150}
  \end{APACrefURL}
\newblock
\APACrefnote{arXiv:2004.05150 [cs]}
\PrintBackRefs{\CurrentBib}

\bibitem [\protect \citeauthoryear {%
Bhaskar%
, Fabbri%
\BCBL {}\ \BBA {} Durrett%
}{%
Bhaskar%
\ \protect \BOthers {.}}{%
{\protect \APACyear {2023}}%
}]{%
bhaskar_prompted_2023}
\APACinsertmetastar {%
bhaskar_prompted_2023}%
\begin{APACrefauthors}%
Bhaskar, A.%
, Fabbri, A.%
\BCBL {}\ \BBA {} Durrett, G.%
\end{APACrefauthors}%
\unskip\
\newblock
\APACrefYearMonthDay{2023}{{\APACmonth{07}}}{}.
\newblock
{\BBOQ}\APACrefatitle {Prompted {Opinion} {Summarization} with {GPT}-3.5}
  {Prompted {Opinion} {Summarization} with {GPT}-3.5}.{\BBCQ}
\newblock
\BIn{} \APACrefbtitle {Findings of the {Association} for {Computational}
  {Linguistics}: {ACL} 2023} {Findings of the {Association} for {Computational}
  {Linguistics}: {ACL} 2023}\ (\BPGS\ 9282--9300).
\newblock
\APACaddressPublisher{Toronto, Canada}{Association for Computational
  Linguistics}.
\newblock
\begin{APACrefURL}
  [{2023-07-20}]\url{https://aclanthology.org/2023.findings-acl.591}
  \end{APACrefURL}
\PrintBackRefs{\CurrentBib}

\bibitem [\protect \citeauthoryear {%
{BigScience Workshop}%
}{%
{BigScience Workshop}%
}{%
{\protect \APACyear {2022}}%
}]{%
bigscience_workshop_bloom_2022}
\APACinsertmetastar {%
bigscience_workshop_bloom_2022}%
\begin{APACrefauthors}%
{BigScience Workshop}.%
\end{APACrefauthors}%
\unskip\
\newblock
\APACrefYearMonthDay{2022}{{\APACmonth{12}}}{}.
\newblock
\APACrefbtitle {{BLOOM}: {A} {176B}-{Parameter} {Open}-{Access} {Multilingual}
  {Language} {Model}.} {{BLOOM}: {A} {176B}-{Parameter} {Open}-{Access}
  {Multilingual} {Language} {Model}.}
\newblock
\APACaddressPublisher{}{arXiv}.
\newblock
\begin{APACrefURL} [{2023-02-28}]\url{http://arxiv.org/abs/2211.05100}
  \end{APACrefURL}
\newblock
\APACrefnote{arXiv:2211.05100 [cs]}
\newblock
\begin{APACrefDOI} \doi{10.48550/arXiv.2211.05100} \end{APACrefDOI}
\PrintBackRefs{\CurrentBib}

\bibitem [\protect \citeauthoryear {%
Brown%
\ \protect \BOthers {.}}{%
Brown%
\ \protect \BOthers {.}}{%
{\protect \APACyear {2020}}%
}]{%
brown_language_2020}
\APACinsertmetastar {%
brown_language_2020}%
\begin{APACrefauthors}%
Brown, T.%
, Mann, B.%
, Ryder, N.%
, Subbiah, M.%
, Kaplan, J\BPBI D.%
, Dhariwal, P.%
\BDBL {}Amodei, D.%
\end{APACrefauthors}%
\unskip\
\newblock
\APACrefYearMonthDay{2020}{}{}.
\newblock
{\BBOQ}\APACrefatitle {Language {Models} are {Few}-{Shot} {Learners}} {Language
  {Models} are {Few}-{Shot} {Learners}}.{\BBCQ}
\newblock
\BIn{} H.~Larochelle, M.~Ranzato, R.~Hadsell, M\BPBI F.~Balcan\BCBL {}\ \BBA {}
  H.~Lin\ (\BEDS), \APACrefbtitle {Advances in {Neural} {Information}
  {Processing} {Systems}} {Advances in {Neural} {Information} {Processing}
  {Systems}}\ (\BVOL~33, \BPGS\ 1877--1901).
\newblock
\APACaddressPublisher{}{Curran Associates, Inc.}
\newblock
\begin{APACrefURL}
  \url{https://proceedings.neurips.cc/paper_files/paper/2020/file/1457c0d6bfcb4967418bfb8ac142f64a-Paper.pdf}
  \end{APACrefURL}
\PrintBackRefs{\CurrentBib}

\bibitem [\protect \citeauthoryear {%
Butcher%
}{%
Butcher%
}{%
{\protect \APACyear {2022}}%
}]{%
butcher_heres_2022}
\APACinsertmetastar {%
butcher_heres_2022}%
\begin{APACrefauthors}%
Butcher, M.%
\end{APACrefauthors}%
\unskip\
\newblock
\APACrefYearMonthDay{2022}{{\APACmonth{07}}}{}.
\newblock
\APACrefbtitle {Here's why a gold rush of {NLP} startups is about to arrive.}
  {Here's why a gold rush of {NLP} startups is about to arrive.}
\newblock
\begin{APACrefURL}
  [{2023-03-08}]\url{https://techcrunch.com/2022/07/28/a-gold-rush-of-nlp-startups-is-about-to-arrive-heres-why/}
  \end{APACrefURL}
\PrintBackRefs{\CurrentBib}

\bibitem [\protect \citeauthoryear {%
Cachola%
, Lo%
, Cohan%
\BCBL {}\ \BBA {} Weld%
}{%
Cachola%
\ \protect \BOthers {.}}{%
{\protect \APACyear {2020}}%
}]{%
cachola_tldr_2020}
\APACinsertmetastar {%
cachola_tldr_2020}%
\begin{APACrefauthors}%
Cachola, I.%
, Lo, K.%
, Cohan, A.%
\BCBL {}\ \BBA {} Weld, D.%
\end{APACrefauthors}%
\unskip\
\newblock
\APACrefYearMonthDay{2020}{{\APACmonth{11}}}{}.
\newblock
{\BBOQ}\APACrefatitle {{TLDR}: {Extreme} {Summarization} of {Scientific}
  {Documents}} {{TLDR}: {Extreme} {Summarization} of {Scientific}
  {Documents}}.{\BBCQ}
\newblock
\BIn{} \APACrefbtitle {Findings of the {Association} for {Computational}
  {Linguistics}: {EMNLP} 2020} {Findings of the {Association} for
  {Computational} {Linguistics}: {EMNLP} 2020}\ (\BPGS\ 4766--4777).
\newblock
\APACaddressPublisher{Online}{Association for Computational Linguistics}.
\newblock
\begin{APACrefURL}
  [{2023-07-20}]\url{https://aclanthology.org/2020.findings-emnlp.428}
  \end{APACrefURL}
\newblock
\begin{APACrefDOI} \doi{10.18653/v1/2020.findings-emnlp.428} \end{APACrefDOI}
\PrintBackRefs{\CurrentBib}

\bibitem [\protect \citeauthoryear {%
Chowdhery%
\ \protect \BOthers {.}}{%
Chowdhery%
\ \protect \BOthers {.}}{%
{\protect \APACyear {2022}}%
}]{%
chowdhery_palm_2022}
\APACinsertmetastar {%
chowdhery_palm_2022}%
\begin{APACrefauthors}%
Chowdhery, A.%
, Narang, S.%
, Devlin, J.%
, Bosma, M.%
, Mishra, G.%
, Roberts, A.%
\BDBL {}Fiedel, N.%
\end{APACrefauthors}%
\unskip\
\newblock
\APACrefYearMonthDay{2022}{{\APACmonth{10}}}{}.
\newblock
\APACrefbtitle {{PaLM}: {Scaling} {Language} {Modeling} with {Pathways}.}
  {{PaLM}: {Scaling} {Language} {Modeling} with {Pathways}.}
\newblock
\APACaddressPublisher{}{arXiv}.
\newblock
\begin{APACrefURL} [{2023-02-28}]\url{http://arxiv.org/abs/2204.02311}
  \end{APACrefURL}
\newblock
\APACrefnote{arXiv:2204.02311 [cs]}
\newblock
\begin{APACrefDOI} \doi{10.48550/arXiv.2204.02311} \end{APACrefDOI}
\PrintBackRefs{\CurrentBib}

\bibitem [\protect \citeauthoryear {%
Christiano%
\ \protect \BOthers {.}}{%
Christiano%
\ \protect \BOthers {.}}{%
{\protect \APACyear {2017}}%
}]{%
christiano_deep_2017}
\APACinsertmetastar {%
christiano_deep_2017}%
\begin{APACrefauthors}%
Christiano, P\BPBI F.%
, Leike, J.%
, Brown, T.%
, Martic, M.%
, Legg, S.%
\BCBL {}\ \BBA {} Amodei, D.%
\end{APACrefauthors}%
\unskip\
\newblock
\APACrefYearMonthDay{2017}{}{}.
\newblock
{\BBOQ}\APACrefatitle {Deep {Reinforcement} {Learning} from {Human}
  {Preferences}} {Deep {Reinforcement} {Learning} from {Human}
  {Preferences}}.{\BBCQ}
\newblock
\BIn{} I.~Guyon\ \BOthers {.}\ (\BEDS), \APACrefbtitle {Advances in {Neural}
  {Information} {Processing} {Systems}} {Advances in {Neural} {Information}
  {Processing} {Systems}}\ (\BVOL~30).
\newblock
\APACaddressPublisher{}{Curran Associates, Inc.}
\newblock
\begin{APACrefURL}
  \url{https://proceedings.neurips.cc/paper_files/paper/2017/file/d5e2c0adad503c91f91df240d0cd4e49-Paper.pdf}
  \end{APACrefURL}
\PrintBackRefs{\CurrentBib}

\bibitem [\protect \citeauthoryear {%
Chung%
\ \protect \BOthers {.}}{%
Chung%
\ \protect \BOthers {.}}{%
{\protect \APACyear {2022}}%
}]{%
chung_scaling_2022}
\APACinsertmetastar {%
chung_scaling_2022}%
\begin{APACrefauthors}%
Chung, H\BPBI W.%
, Hou, L.%
, Longpre, S.%
, Zoph, B.%
, Tay, Y.%
, Fedus, W.%
\BDBL {}Wei, J.%
\end{APACrefauthors}%
\unskip\
\newblock
\APACrefYearMonthDay{2022}{{\APACmonth{12}}}{}.
\newblock
\APACrefbtitle {Scaling {Instruction}-{Finetuned} {Language} {Models}.}
  {Scaling {Instruction}-{Finetuned} {Language} {Models}.}
\newblock
\APACaddressPublisher{}{arXiv}.
\newblock
\begin{APACrefURL} [{2023-03-01}]\url{http://arxiv.org/abs/2210.11416}
  \end{APACrefURL}
\newblock
\APACrefnote{arXiv:2210.11416 [cs]}
\PrintBackRefs{\CurrentBib}

\bibitem [\protect \citeauthoryear {%
Colombo%
, Peyrard%
, Noiry%
, West%
\BCBL {}\ \BBA {} Piantanida%
}{%
Colombo%
\ \protect \BOthers {.}}{%
{\protect \APACyear {2022}}%
}]{%
colombo_glass_2022}
\APACinsertmetastar {%
colombo_glass_2022}%
\begin{APACrefauthors}%
Colombo, P.%
, Peyrard, M.%
, Noiry, N.%
, West, R.%
\BCBL {}\ \BBA {} Piantanida, P.%
\end{APACrefauthors}%
\unskip\
\newblock
\APACrefYearMonthDay{2022}{{\APACmonth{10}}}{}.
\newblock
\APACrefbtitle {The {Glass} {Ceiling} of {Automatic} {Evaluation} in {Natural}
  {Language} {Generation}.} {The {Glass} {Ceiling} of {Automatic} {Evaluation}
  in {Natural} {Language} {Generation}.}
\newblock
\APACaddressPublisher{}{arXiv}.
\newblock
\begin{APACrefURL} [{2023-03-16}]\url{http://arxiv.org/abs/2208.14585}
  \end{APACrefURL}
\newblock
\APACrefnote{arXiv:2208.14585 [cs] version: 2}
\PrintBackRefs{\CurrentBib}

\bibitem [\protect \citeauthoryear {%
Fabbri%
\ \protect \BOthers {.}}{%
Fabbri%
\ \protect \BOthers {.}}{%
{\protect \APACyear {2021}}%
}]{%
fabbri_summeval_2021}
\APACinsertmetastar {%
fabbri_summeval_2021}%
\begin{APACrefauthors}%
Fabbri, A\BPBI R.%
, Kryściński, W.%
, McCann, B.%
, Xiong, C.%
, Socher, R.%
\BCBL {}\ \BBA {} Radev, D.%
\end{APACrefauthors}%
\unskip\
\newblock
\APACrefYearMonthDay{2021}{{\APACmonth{04}}}{}.
\newblock
{\BBOQ}\APACrefatitle {{SummEval}: {Re}-evaluating {Summarization}
  {Evaluation}} {{SummEval}: {Re}-evaluating {Summarization}
  {Evaluation}}.{\BBCQ}
\newblock
\APACjournalVolNumPages{Transactions of the Association for Computational
  Linguistics}{9}{}{391--409}.
\newblock
\begin{APACrefURL} [{2023-07-20}]\url{https://doi.org/10.1162/tacl_a_00373}
  \end{APACrefURL}
\newblock
\begin{APACrefDOI} \doi{10.1162/tacl_a_00373} \end{APACrefDOI}
\PrintBackRefs{\CurrentBib}

\bibitem [\protect \citeauthoryear {%
Gidiotis%
\ \BBA {} Tsoumakas%
}{%
Gidiotis%
\ \BBA {} Tsoumakas%
}{%
{\protect \APACyear {2020}}%
}]{%
gidiotis_divide-and-conquer_2020}
\APACinsertmetastar {%
gidiotis_divide-and-conquer_2020}%
\begin{APACrefauthors}%
Gidiotis, A.%
\BCBT {}\ \BBA {} Tsoumakas, G.%
\end{APACrefauthors}%
\unskip\
\newblock
\APACrefYearMonthDay{2020}{}{}.
\newblock
{\BBOQ}\APACrefatitle {A {Divide}-and-{Conquer} {Approach} to the
  {Summarization} of {Long} {Documents}} {A {Divide}-and-{Conquer} {Approach}
  to the {Summarization} of {Long} {Documents}}.{\BBCQ}
\newblock
\APACjournalVolNumPages{IEEE/ACM Transactions on Audio, Speech, and Language
  Processing}{28}{}{3029--3040}.
\newblock
\APACrefnote{Conference Name: IEEE/ACM Transactions on Audio, Speech, and
  Language Processing}
\newblock
\begin{APACrefDOI} \doi{10.1109/TASLP.2020.3037401} \end{APACrefDOI}
\PrintBackRefs{\CurrentBib}

\bibitem [\protect \citeauthoryear {%
Gliwa%
, Mochol%
, Biesek%
\BCBL {}\ \BBA {} Wawer%
}{%
Gliwa%
\ \protect \BOthers {.}}{%
{\protect \APACyear {2019}}%
}]{%
gliwa_samsum_2019}
\APACinsertmetastar {%
gliwa_samsum_2019}%
\begin{APACrefauthors}%
Gliwa, B.%
, Mochol, I.%
, Biesek, M.%
\BCBL {}\ \BBA {} Wawer, A.%
\end{APACrefauthors}%
\unskip\
\newblock
\APACrefYearMonthDay{2019}{{\APACmonth{11}}}{}.
\newblock
{\BBOQ}\APACrefatitle {{SAMSum} {Corpus}: {A} {Human}-annotated {Dialogue}
  {Dataset} for {Abstractive} {Summarization}} {{SAMSum} {Corpus}: {A}
  {Human}-annotated {Dialogue} {Dataset} for {Abstractive}
  {Summarization}}.{\BBCQ}
\newblock
\BIn{} \APACrefbtitle {Proceedings of the 2nd {Workshop} on {New} {Frontiers}
  in {Summarization}} {Proceedings of the 2nd {Workshop} on {New} {Frontiers}
  in {Summarization}}\ (\BPGS\ 70--79).
\newblock
\APACaddressPublisher{Hong Kong, China}{Association for Computational
  Linguistics}.
\newblock
\begin{APACrefURL} [{2023-03-01}]\url{https://aclanthology.org/D19-5409}
  \end{APACrefURL}
\newblock
\begin{APACrefDOI} \doi{10.18653/v1/D19-5409} \end{APACrefDOI}
\PrintBackRefs{\CurrentBib}

\bibitem [\protect \citeauthoryear {%
Goodwin%
, Savery%
\BCBL {}\ \BBA {} Demner-Fushman%
}{%
Goodwin%
\ \protect \BOthers {.}}{%
{\protect \APACyear {2020}}%
}]{%
goodwin_flight_2020}
\APACinsertmetastar {%
goodwin_flight_2020}%
\begin{APACrefauthors}%
Goodwin, T.%
, Savery, M.%
\BCBL {}\ \BBA {} Demner-Fushman, D.%
\end{APACrefauthors}%
\unskip\
\newblock
\APACrefYearMonthDay{2020}{{\APACmonth{12}}}{}.
\newblock
{\BBOQ}\APACrefatitle {Flight of the {PEGASUS}? {Comparing} {Transformers} on
  {Few}-shot and {Zero}-shot {Multi}-document {Abstractive} {Summarization}}
  {Flight of the {PEGASUS}? {Comparing} {Transformers} on {Few}-shot and
  {Zero}-shot {Multi}-document {Abstractive} {Summarization}}.{\BBCQ}
\newblock
\BIn{} \APACrefbtitle {Proceedings of the 28th {International} {Conference} on
  {Computational} {Linguistics}} {Proceedings of the 28th {International}
  {Conference} on {Computational} {Linguistics}}\ (\BPGS\ 5640--5646).
\newblock
\APACaddressPublisher{Barcelona, Spain (Online)}{International Committee on
  Computational Linguistics}.
\newblock
\begin{APACrefURL}
  [{2023-05-07}]\url{https://aclanthology.org/2020.coling-main.494}
  \end{APACrefURL}
\newblock
\begin{APACrefDOI} \doi{10.18653/v1/2020.coling-main.494} \end{APACrefDOI}
\PrintBackRefs{\CurrentBib}

\bibitem [\protect \citeauthoryear {%
Goyal%
, Li%
\BCBL {}\ \BBA {} Durrett%
}{%
Goyal%
\ \protect \BOthers {.}}{%
{\protect \APACyear {2022}}%
}]{%
goyal_news_2022}
\APACinsertmetastar {%
goyal_news_2022}%
\begin{APACrefauthors}%
Goyal, T.%
, Li, J\BPBI J.%
\BCBL {}\ \BBA {} Durrett, G.%
\end{APACrefauthors}%
\unskip\
\newblock
\APACrefYearMonthDay{2022}{{\APACmonth{09}}}{}.
\newblock
\APACrefbtitle {News {Summarization} and {Evaluation} in the {Era} of {GPT}-3.}
  {News {Summarization} and {Evaluation} in the {Era} of {GPT}-3.}
\newblock
\APACaddressPublisher{}{arXiv}.
\newblock
\begin{APACrefURL} [{2023-02-17}]\url{http://arxiv.org/abs/2209.12356}
  \end{APACrefURL}
\newblock
\APACrefnote{arXiv:2209.12356 [cs]}
\PrintBackRefs{\CurrentBib}

\bibitem [\protect \citeauthoryear {%
Guo%
\ \protect \BOthers {.}}{%
Guo%
\ \protect \BOthers {.}}{%
{\protect \APACyear {2022}}%
}]{%
guo_longt5_2022}
\APACinsertmetastar {%
guo_longt5_2022}%
\begin{APACrefauthors}%
Guo, M.%
, Ainslie, J.%
, Uthus, D.%
, Ontanon, S.%
, Ni, J.%
, Sung, Y\BHBI H.%
\BCBL {}\ \BBA {} Yang, Y.%
\end{APACrefauthors}%
\unskip\
\newblock
\APACrefYearMonthDay{2022}{{\APACmonth{07}}}{}.
\newblock
{\BBOQ}\APACrefatitle {{LongT5}: {Efficient} {Text}-{To}-{Text} {Transformer}
  for {Long} {Sequences}} {{LongT5}: {Efficient} {Text}-{To}-{Text}
  {Transformer} for {Long} {Sequences}}.{\BBCQ}
\newblock
\BIn{} \APACrefbtitle {Findings of the {Association} for {Computational}
  {Linguistics}: {NAACL} 2022} {Findings of the {Association} for
  {Computational} {Linguistics}: {NAACL} 2022}\ (\BPGS\ 724--736).
\newblock
\APACaddressPublisher{Seattle, United States}{Association for Computational
  Linguistics}.
\newblock
\begin{APACrefURL}
  [{2023-07-20}]\url{https://aclanthology.org/2022.findings-naacl.55}
  \end{APACrefURL}
\newblock
\begin{APACrefDOI} \doi{10.18653/v1/2022.findings-naacl.55} \end{APACrefDOI}
\PrintBackRefs{\CurrentBib}

\bibitem [\protect \citeauthoryear {%
J.~He%
, Kryscinski%
, McCann%
, Rajani%
\BCBL {}\ \BBA {} Xiong%
}{%
J.~He%
\ \protect \BOthers {.}}{%
{\protect \APACyear {2022}}%
}]{%
he_ctrlsum_2022}
\APACinsertmetastar {%
he_ctrlsum_2022}%
\begin{APACrefauthors}%
He, J.%
, Kryscinski, W.%
, McCann, B.%
, Rajani, N.%
\BCBL {}\ \BBA {} Xiong, C.%
\end{APACrefauthors}%
\unskip\
\newblock
\APACrefYearMonthDay{2022}{{\APACmonth{12}}}{}.
\newblock
{\BBOQ}\APACrefatitle {{CTRLsum}: {Towards} {Generic} {Controllable} {Text}
  {Summarization}} {{CTRLsum}: {Towards} {Generic} {Controllable} {Text}
  {Summarization}}.{\BBCQ}
\newblock
\BIn{} \APACrefbtitle {Proceedings of the 2022 {Conference} on {Empirical}
  {Methods} in {Natural} {Language} {Processing}} {Proceedings of the 2022
  {Conference} on {Empirical} {Methods} in {Natural} {Language} {Processing}}\
  (\BPGS\ 5879--5915).
\newblock
\APACaddressPublisher{Abu Dhabi, United Arab Emirates}{Association for
  Computational Linguistics}.
\newblock
\begin{APACrefURL}
  [{2023-07-21}]\url{https://aclanthology.org/2022.emnlp-main.396}
  \end{APACrefURL}
\PrintBackRefs{\CurrentBib}

\bibitem [\protect \citeauthoryear {%
P.~He%
\ \protect \BOthers {.}}{%
P.~He%
\ \protect \BOthers {.}}{%
{\protect \APACyear {2023}}%
}]{%
he_z-code_2023}
\APACinsertmetastar {%
he_z-code_2023}%
\begin{APACrefauthors}%
He, P.%
, Peng, B.%
, Wang, S.%
, Liu, Y.%
, Xu, R.%
, Hassan, H.%
\BDBL {}Huang, X.%
\end{APACrefauthors}%
\unskip\
\newblock
\APACrefYearMonthDay{2023}{{\APACmonth{07}}}{}.
\newblock
{\BBOQ}\APACrefatitle {Z-{Code}++: {A} {Pre}-trained {Language} {Model}
  {Optimized} for {Abstractive} {Summarization}} {Z-{Code}++: {A} {Pre}-trained
  {Language} {Model} {Optimized} for {Abstractive} {Summarization}}.{\BBCQ}
\newblock
\BIn{} \APACrefbtitle {Proceedings of the 61st {Annual} {Meeting} of the
  {Association} for {Computational} {Linguistics} ({Volume} 1: {Long}
  {Papers})} {Proceedings of the 61st {Annual} {Meeting} of the {Association}
  for {Computational} {Linguistics} ({Volume} 1: {Long} {Papers})}\ (\BPGS\
  5095--5112).
\newblock
\APACaddressPublisher{Toronto, Canada}{Association for Computational
  Linguistics}.
\newblock
\begin{APACrefURL}
  [{2023-07-20}]\url{https://aclanthology.org/2023.acl-long.279}
  \end{APACrefURL}
\PrintBackRefs{\CurrentBib}

\bibitem [\protect \citeauthoryear {%
Hendrycks%
\ \protect \BOthers {.}}{%
Hendrycks%
\ \protect \BOthers {.}}{%
{\protect \APACyear {2020}}%
}]{%
hendrycks_measuring_2020}
\APACinsertmetastar {%
hendrycks_measuring_2020}%
\begin{APACrefauthors}%
Hendrycks, D.%
, Burns, C.%
, Basart, S.%
, Zou, A.%
, Mazeika, M.%
, Song, D.%
\BCBL {}\ \BBA {} Steinhardt, J.%
\end{APACrefauthors}%
\unskip\
\newblock
\APACrefYearMonthDay{2020}{{\APACmonth{10}}}{}.
\newblock
{\BBOQ}\APACrefatitle {Measuring {Massive} {Multitask} {Language}
  {Understanding}} {Measuring {Massive} {Multitask} {Language}
  {Understanding}}.{\BBCQ}
\newblock
\BIn{} \APACrefbtitle {9th {International} {Conference} on {Learning}
  {Representations}, {ICLR} 2021, {Virtual} {Event}, {Austria}, {May} 3-7,
  2021.} {9th {International} {Conference} on {Learning} {Representations},
  {ICLR} 2021, {Virtual} {Event}, {Austria}, {May} 3-7, 2021.}
\newblock
\begin{APACrefURL}
  [{2023-07-22}]\url{https://openreview.net/forum?id=d7KBjmI3GmQ}
  \end{APACrefURL}
\PrintBackRefs{\CurrentBib}

\bibitem [\protect \citeauthoryear {%
Herskowitz%
}{%
Herskowitz%
}{%
{\protect \APACyear {2023}}%
}]{%
herskowitz_microsoft_2023}
\APACinsertmetastar {%
herskowitz_microsoft_2023}%
\begin{APACrefauthors}%
Herskowitz, N.%
\end{APACrefauthors}%
\unskip\
\newblock
\APACrefYearMonthDay{2023}{{\APACmonth{02}}}{}.
\newblock
\APACrefbtitle {Microsoft {Teams} {Premium}: {Cut} costs and add {AI}-powered
  productivity.} {Microsoft {Teams} {Premium}: {Cut} costs and add {AI}-powered
  productivity.}
\newblock
\begin{APACrefURL}
  [{2023-02-28}]\url{https://www.microsoft.com/en-us/microsoft-365/blog/2023/02/01/microsoft-teams-premium-cut-costs-and-add-ai-powered-productivity/}
  \end{APACrefURL}
\PrintBackRefs{\CurrentBib}

\bibitem [\protect \citeauthoryear {%
Hillier%
\ \BBA {} Gallé%
}{%
Hillier%
\ \BBA {} Gallé%
}{%
{\protect \APACyear {2023}}%
}]{%
hillier_introducing_2023}
\APACinsertmetastar {%
hillier_introducing_2023}%
\begin{APACrefauthors}%
Hillier, S.%
\BCBT {}\ \BBA {} Gallé, M.%
\end{APACrefauthors}%
\unskip\
\newblock
\APACrefYearMonthDay{2023}{{\APACmonth{02}}}{}.
\newblock
\APACrefbtitle {Introducing {Cohere} {Summarize} {Beta}: {A} {New} {Endpoint}
  for {Text} {Summarization}.} {Introducing {Cohere} {Summarize} {Beta}: {A}
  {New} {Endpoint} for {Text} {Summarization}.}
\newblock
\begin{APACrefURL} [{2023-03-01}]\url{https://txt.cohere.ai/summarize-beta/}
  \end{APACrefURL}
\PrintBackRefs{\CurrentBib}

\bibitem [\protect \citeauthoryear {%
Hoffmann%
\ \protect \BOthers {.}}{%
Hoffmann%
\ \protect \BOthers {.}}{%
{\protect \APACyear {2022}}%
}]{%
hoffmann_training_2022}
\APACinsertmetastar {%
hoffmann_training_2022}%
\begin{APACrefauthors}%
Hoffmann, J.%
, Borgeaud, S.%
, Mensch, A.%
, Buchatskaya, E.%
, Cai, T.%
, Rutherford, E.%
\BDBL {}Sifre, L.%
\end{APACrefauthors}%
\unskip\
\newblock
\APACrefYearMonthDay{2022}{{\APACmonth{03}}}{}.
\newblock
\APACrefbtitle {Training {Compute}-{Optimal} {Large} {Language} {Models}.}
  {Training {Compute}-{Optimal} {Large} {Language} {Models}.}
\newblock
\APACaddressPublisher{}{arXiv}.
\newblock
\begin{APACrefURL} [{2023-02-28}]\url{http://arxiv.org/abs/2203.15556}
  \end{APACrefURL}
\newblock
\APACrefnote{arXiv:2203.15556 [cs]}
\PrintBackRefs{\CurrentBib}

\bibitem [\protect \citeauthoryear {%
Iyer%
\ \protect \BOthers {.}}{%
Iyer%
\ \protect \BOthers {.}}{%
{\protect \APACyear {2023}}%
}]{%
iyer_opt-iml_2023}
\APACinsertmetastar {%
iyer_opt-iml_2023}%
\begin{APACrefauthors}%
Iyer, S.%
, Lin, X\BPBI V.%
, Pasunuru, R.%
, Mihaylov, T.%
, Simig, D.%
, Yu, P.%
\BDBL {}Stoyanov, V.%
\end{APACrefauthors}%
\unskip\
\newblock
\APACrefYearMonthDay{2023}{{\APACmonth{01}}}{}.
\newblock
\APACrefbtitle {{OPT}-{IML}: {Scaling} {Language} {Model} {Instruction} {Meta}
  {Learning} through the {Lens} of {Generalization}.} {{OPT}-{IML}: {Scaling}
  {Language} {Model} {Instruction} {Meta} {Learning} through the {Lens} of
  {Generalization}.}
\newblock
\APACaddressPublisher{}{arXiv}.
\newblock
\begin{APACrefURL} [{2023-03-01}]\url{http://arxiv.org/abs/2212.12017}
  \end{APACrefURL}
\newblock
\APACrefnote{arXiv:2212.12017 [cs]}
\newblock
\begin{APACrefDOI} \doi{10.48550/arXiv.2212.12017} \end{APACrefDOI}
\PrintBackRefs{\CurrentBib}

\bibitem [\protect \citeauthoryear {%
Jaegle%
\ \protect \BOthers {.}}{%
Jaegle%
\ \protect \BOthers {.}}{%
{\protect \APACyear {2021}}%
}]{%
jaegle_perceiver_2021}
\APACinsertmetastar {%
jaegle_perceiver_2021}%
\begin{APACrefauthors}%
Jaegle, A.%
, Borgeaud, S.%
, Alayrac, J\BHBI B.%
, Doersch, C.%
, Ionescu, C.%
, Ding, D.%
\BDBL {}Carreira, J.%
\end{APACrefauthors}%
\unskip\
\newblock
\APACrefYearMonthDay{2021}{{\APACmonth{10}}}{}.
\newblock
{\BBOQ}\APACrefatitle {Perceiver {IO}: {A} {General} {Architecture} for
  {Structured} {Inputs} \& {Outputs}} {Perceiver {IO}: {A} {General}
  {Architecture} for {Structured} {Inputs} \& {Outputs}}.{\BBCQ}
\newblock
\BIn{} \APACrefbtitle {The {Tenth} {International} {Conference} on {Learning}
  {Representations}, {ICLR} 2022, {Virtual} {Event}, {April} 25-29, 2022.} {The
  {Tenth} {International} {Conference} on {Learning} {Representations}, {ICLR}
  2022, {Virtual} {Event}, {April} 25-29, 2022.}
\newblock
\begin{APACrefURL}
  [{2023-07-22}]\url{https://openreview.net/forum?id=fILj7WpI-g}
  \end{APACrefURL}
\PrintBackRefs{\CurrentBib}

\bibitem [\protect \citeauthoryear {%
Jangra%
, Mukherjee%
, Jatowt%
, Saha%
\BCBL {}\ \BBA {} Hasanuzzaman%
}{%
Jangra%
\ \protect \BOthers {.}}{%
{\protect \APACyear {2023}}%
}]{%
jangra_survey_2023}
\APACinsertmetastar {%
jangra_survey_2023}%
\begin{APACrefauthors}%
Jangra, A.%
, Mukherjee, S.%
, Jatowt, A.%
, Saha, S.%
\BCBL {}\ \BBA {} Hasanuzzaman, M.%
\end{APACrefauthors}%
\unskip\
\newblock
\APACrefYearMonthDay{2023}{{\APACmonth{02}}}{}.
\newblock
\APACrefbtitle {A {Survey} on {Multi}-modal {Summarization}.} {A {Survey} on
  {Multi}-modal {Summarization}.}
\newblock
\APACaddressPublisher{}{arXiv}.
\newblock
\begin{APACrefURL} [{2023-05-01}]\url{http://arxiv.org/abs/2109.05199}
  \end{APACrefURL}
\newblock
\APACrefnote{arXiv:2109.05199 [cs]}
\PrintBackRefs{\CurrentBib}

\bibitem [\protect \citeauthoryear {%
Ji%
\ \protect \BOthers {.}}{%
Ji%
\ \protect \BOthers {.}}{%
{\protect \APACyear {2023}}%
}]{%
ji_survey_2023}
\APACinsertmetastar {%
ji_survey_2023}%
\begin{APACrefauthors}%
Ji, Z.%
, Lee, N.%
, Frieske, R.%
, Yu, T.%
, Su, D.%
, Xu, Y.%
\BDBL {}Fung, P.%
\end{APACrefauthors}%
\unskip\
\newblock
\APACrefYearMonthDay{2023}{{\APACmonth{12}}}{}.
\newblock
{\BBOQ}\APACrefatitle {Survey of {Hallucination} in {Natural} {Language}
  {Generation}} {Survey of {Hallucination} in {Natural} {Language}
  {Generation}}.{\BBCQ}
\newblock
\APACjournalVolNumPages{ACM Computing Surveys}{55}{12}{1--38}.
\newblock
\begin{APACrefURL} [{2023-03-06}]\url{http://arxiv.org/abs/2202.03629}
  \end{APACrefURL}
\newblock
\APACrefnote{arXiv:2202.03629 [cs]}
\newblock
\begin{APACrefDOI} \doi{10.1145/3571730} \end{APACrefDOI}
\PrintBackRefs{\CurrentBib}

\bibitem [\protect \citeauthoryear {%
Kadavath%
\ \protect \BOthers {.}}{%
Kadavath%
\ \protect \BOthers {.}}{%
{\protect \APACyear {2022}}%
}]{%
kadavath_language_2022}
\APACinsertmetastar {%
kadavath_language_2022}%
\begin{APACrefauthors}%
Kadavath, S.%
, Conerly, T.%
, Askell, A.%
, Henighan, T.%
, Drain, D.%
, Perez, E.%
\BDBL {}Kaplan, J.%
\end{APACrefauthors}%
\unskip\
\newblock
\APACrefYearMonthDay{2022}{{\APACmonth{11}}}{}.
\newblock
\APACrefbtitle {Language {Models} ({Mostly}) {Know} {What} {They} {Know}.}
  {Language {Models} ({Mostly}) {Know} {What} {They} {Know}.}
\newblock
\APACaddressPublisher{}{arXiv}.
\newblock
\begin{APACrefURL} [{2023-03-02}]\url{http://arxiv.org/abs/2207.05221}
  \end{APACrefURL}
\newblock
\APACrefnote{arXiv:2207.05221 [cs]}
\newblock
\begin{APACrefDOI} \doi{10.48550/arXiv.2207.05221} \end{APACrefDOI}
\PrintBackRefs{\CurrentBib}

\bibitem [\protect \citeauthoryear {%
Karpukhin%
\ \protect \BOthers {.}}{%
Karpukhin%
\ \protect \BOthers {.}}{%
{\protect \APACyear {2020}}%
}]{%
karpukhin_dense_2020}
\APACinsertmetastar {%
karpukhin_dense_2020}%
\begin{APACrefauthors}%
Karpukhin, V.%
, Oguz, B.%
, Min, S.%
, Lewis, P.%
, Wu, L.%
, Edunov, S.%
\BDBL {}Yih, W\BHBI t.%
\end{APACrefauthors}%
\unskip\
\newblock
\APACrefYearMonthDay{2020}{{\APACmonth{11}}}{}.
\newblock
{\BBOQ}\APACrefatitle {Dense {Passage} {Retrieval} for {Open}-{Domain}
  {Question} {Answering}} {Dense {Passage} {Retrieval} for {Open}-{Domain}
  {Question} {Answering}}.{\BBCQ}
\newblock
\BIn{} \APACrefbtitle {Proceedings of the 2020 {Conference} on {Empirical}
  {Methods} in {Natural} {Language} {Processing} ({EMNLP})} {Proceedings of the
  2020 {Conference} on {Empirical} {Methods} in {Natural} {Language}
  {Processing} ({EMNLP})}\ (\BPGS\ 6769--6781).
\newblock
\APACaddressPublisher{Online}{Association for Computational Linguistics}.
\newblock
\begin{APACrefURL}
  [{2023-07-21}]\url{https://aclanthology.org/2020.emnlp-main.550}
  \end{APACrefURL}
\newblock
\begin{APACrefDOI} \doi{10.18653/v1/2020.emnlp-main.550} \end{APACrefDOI}
\PrintBackRefs{\CurrentBib}

\bibitem [\protect \citeauthoryear {%
Kikuchi%
, Neubig%
, Sasano%
, Takamura%
\BCBL {}\ \BBA {} Okumura%
}{%
Kikuchi%
\ \protect \BOthers {.}}{%
{\protect \APACyear {2016}}%
}]{%
kikuchi_controlling_2016}
\APACinsertmetastar {%
kikuchi_controlling_2016}%
\begin{APACrefauthors}%
Kikuchi, Y.%
, Neubig, G.%
, Sasano, R.%
, Takamura, H.%
\BCBL {}\ \BBA {} Okumura, M.%
\end{APACrefauthors}%
\unskip\
\newblock
\APACrefYearMonthDay{2016}{{\APACmonth{11}}}{}.
\newblock
{\BBOQ}\APACrefatitle {Controlling {Output} {Length} in {Neural}
  {Encoder}-{Decoders}} {Controlling {Output} {Length} in {Neural}
  {Encoder}-{Decoders}}.{\BBCQ}
\newblock
\BIn{} \APACrefbtitle {Proceedings of the 2016 {Conference} on {Empirical}
  {Methods} in {Natural} {Language} {Processing}} {Proceedings of the 2016
  {Conference} on {Empirical} {Methods} in {Natural} {Language} {Processing}}\
  (\BPGS\ 1328--1338).
\newblock
\APACaddressPublisher{Austin, Texas}{Association for Computational
  Linguistics}.
\newblock
\begin{APACrefURL} [{2023-07-21}]\url{https://aclanthology.org/D16-1140}
  \end{APACrefURL}
\newblock
\begin{APACrefDOI} \doi{10.18653/v1/D16-1140} \end{APACrefDOI}
\PrintBackRefs{\CurrentBib}

\bibitem [\protect \citeauthoryear {%
Kocmi%
\ \BBA {} Federmann%
}{%
Kocmi%
\ \BBA {} Federmann%
}{%
{\protect \APACyear {2023}}%
}]{%
kocmi_large_2023}
\APACinsertmetastar {%
kocmi_large_2023}%
\begin{APACrefauthors}%
Kocmi, T.%
\BCBT {}\ \BBA {} Federmann, C.%
\end{APACrefauthors}%
\unskip\
\newblock
\APACrefYearMonthDay{2023}{{\APACmonth{02}}}{}.
\newblock
\APACrefbtitle {Large {Language} {Models} {Are} {State}-of-the-{Art}
  {Evaluators} of {Translation} {Quality}.} {Large {Language} {Models} {Are}
  {State}-of-the-{Art} {Evaluators} of {Translation} {Quality}.}
\newblock
\APACaddressPublisher{}{arXiv}.
\newblock
\begin{APACrefURL} [{2023-03-02}]\url{http://arxiv.org/abs/2302.14520}
  \end{APACrefURL}
\newblock
\APACrefnote{arXiv:2302.14520 [cs]}
\newblock
\begin{APACrefDOI} \doi{10.48550/arXiv.2302.14520} \end{APACrefDOI}
\PrintBackRefs{\CurrentBib}

\bibitem [\protect \citeauthoryear {%
Kryscinski%
, McCann%
, Xiong%
\BCBL {}\ \BBA {} Socher%
}{%
Kryscinski%
\ \protect \BOthers {.}}{%
{\protect \APACyear {2020}}%
}]{%
kryscinski_evaluating_2020}
\APACinsertmetastar {%
kryscinski_evaluating_2020}%
\begin{APACrefauthors}%
Kryscinski, W.%
, McCann, B.%
, Xiong, C.%
\BCBL {}\ \BBA {} Socher, R.%
\end{APACrefauthors}%
\unskip\
\newblock
\APACrefYearMonthDay{2020}{{\APACmonth{11}}}{}.
\newblock
{\BBOQ}\APACrefatitle {Evaluating the {Factual} {Consistency} of {Abstractive}
  {Text} {Summarization}} {Evaluating the {Factual} {Consistency} of
  {Abstractive} {Text} {Summarization}}.{\BBCQ}
\newblock
\BIn{} \APACrefbtitle {Proceedings of the 2020 {Conference} on {Empirical}
  {Methods} in {Natural} {Language} {Processing} ({EMNLP})} {Proceedings of the
  2020 {Conference} on {Empirical} {Methods} in {Natural} {Language}
  {Processing} ({EMNLP})}\ (\BPGS\ 9332--9346).
\newblock
\APACaddressPublisher{Online}{Association for Computational Linguistics}.
\newblock
\begin{APACrefURL}
  [{2023-07-20}]\url{https://aclanthology.org/2020.emnlp-main.750}
  \end{APACrefURL}
\newblock
\begin{APACrefDOI} \doi{10.18653/v1/2020.emnlp-main.750} \end{APACrefDOI}
\PrintBackRefs{\CurrentBib}

\bibitem [\protect \citeauthoryear {%
Ladhak%
, Li%
, Al-Onaizan%
\BCBL {}\ \BBA {} McKeown%
}{%
Ladhak%
\ \protect \BOthers {.}}{%
{\protect \APACyear {2020}}%
}]{%
ladhak_exploring_2020}
\APACinsertmetastar {%
ladhak_exploring_2020}%
\begin{APACrefauthors}%
Ladhak, F.%
, Li, B.%
, Al-Onaizan, Y.%
\BCBL {}\ \BBA {} McKeown, K.%
\end{APACrefauthors}%
\unskip\
\newblock
\APACrefYearMonthDay{2020}{{\APACmonth{07}}}{}.
\newblock
{\BBOQ}\APACrefatitle {Exploring {Content} {Selection} in {Summarization} of
  {Novel} {Chapters}} {Exploring {Content} {Selection} in {Summarization} of
  {Novel} {Chapters}}.{\BBCQ}
\newblock
\BIn{} \APACrefbtitle {Proceedings of the 58th {Annual} {Meeting} of the
  {Association} for {Computational} {Linguistics}} {Proceedings of the 58th
  {Annual} {Meeting} of the {Association} for {Computational} {Linguistics}}\
  (\BPGS\ 5043--5054).
\newblock
\APACaddressPublisher{Online}{Association for Computational Linguistics}.
\newblock
\begin{APACrefURL}
  [{2023-07-20}]\url{https://aclanthology.org/2020.acl-main.453}
  \end{APACrefURL}
\newblock
\begin{APACrefDOI} \doi{10.18653/v1/2020.acl-main.453} \end{APACrefDOI}
\PrintBackRefs{\CurrentBib}

\bibitem [\protect \citeauthoryear {%
Larkin%
}{%
Larkin%
}{%
{\protect \APACyear {2022}}%
}]{%
larkin_zoom_2022}
\APACinsertmetastar {%
larkin_zoom_2022}%
\begin{APACrefauthors}%
Larkin, T.%
\end{APACrefauthors}%
\unskip\
\newblock
\APACrefYearMonthDay{2022}{{\APACmonth{04}}}{}.
\newblock
\APACrefbtitle {Zoom {IQ} for {Sales}: {Conversational} intelligence for
  sellers.} {Zoom {IQ} for {Sales}: {Conversational} intelligence for sellers.}
\newblock
\begin{APACrefURL} [{2023-07-26}]\url{https://blog.zoom.us/zoom-iq-for-sales/}
  \end{APACrefURL}
\PrintBackRefs{\CurrentBib}

\bibitem [\protect \citeauthoryear {%
Lewis%
\ \protect \BOthers {.}}{%
Lewis%
\ \protect \BOthers {.}}{%
{\protect \APACyear {2020}}%
}]{%
lewis_bart_2020}
\APACinsertmetastar {%
lewis_bart_2020}%
\begin{APACrefauthors}%
Lewis, M.%
, Liu, Y.%
, Goyal, N.%
, Ghazvininejad, M.%
, Mohamed, A.%
, Levy, O.%
\BDBL {}Zettlemoyer, L.%
\end{APACrefauthors}%
\unskip\
\newblock
\APACrefYearMonthDay{2020}{{\APACmonth{07}}}{}.
\newblock
{\BBOQ}\APACrefatitle {{BART}: {Denoising} {Sequence}-to-{Sequence}
  {Pre}-training for {Natural} {Language} {Generation}, {Translation}, and
  {Comprehension}} {{BART}: {Denoising} {Sequence}-to-{Sequence} {Pre}-training
  for {Natural} {Language} {Generation}, {Translation}, and
  {Comprehension}}.{\BBCQ}
\newblock
\BIn{} \APACrefbtitle {Proceedings of the 58th {Annual} {Meeting} of the
  {Association} for {Computational} {Linguistics}} {Proceedings of the 58th
  {Annual} {Meeting} of the {Association} for {Computational} {Linguistics}}\
  (\BPGS\ 7871--7880).
\newblock
\APACaddressPublisher{Online}{Association for Computational Linguistics}.
\newblock
\begin{APACrefURL}
  [{2023-07-21}]\url{https://aclanthology.org/2020.acl-main.703}
  \end{APACrefURL}
\newblock
\begin{APACrefDOI} \doi{10.18653/v1/2020.acl-main.703} \end{APACrefDOI}
\PrintBackRefs{\CurrentBib}

\bibitem [\protect \citeauthoryear {%
H.~Li%
, Zhu%
, Ma%
, Zhang%
\BCBL {}\ \BBA {} Zong%
}{%
H.~Li%
\ \protect \BOthers {.}}{%
{\protect \APACyear {2017}}%
}]{%
li_multi-modal_2017}
\APACinsertmetastar {%
li_multi-modal_2017}%
\begin{APACrefauthors}%
Li, H.%
, Zhu, J.%
, Ma, C.%
, Zhang, J.%
\BCBL {}\ \BBA {} Zong, C.%
\end{APACrefauthors}%
\unskip\
\newblock
\APACrefYearMonthDay{2017}{{\APACmonth{09}}}{}.
\newblock
{\BBOQ}\APACrefatitle {Multi-modal {Summarization} for {Asynchronous}
  {Collection} of {Text}, {Image}, {Audio} and {Video}} {Multi-modal
  {Summarization} for {Asynchronous} {Collection} of {Text}, {Image}, {Audio}
  and {Video}}.{\BBCQ}
\newblock
\BIn{} \APACrefbtitle {Proceedings of the 2017 {Conference} on {Empirical}
  {Methods} in {Natural} {Language} {Processing}} {Proceedings of the 2017
  {Conference} on {Empirical} {Methods} in {Natural} {Language} {Processing}}\
  (\BPGS\ 1092--1102).
\newblock
\APACaddressPublisher{Copenhagen, Denmark}{Association for Computational
  Linguistics}.
\newblock
\begin{APACrefURL} [{2023-05-06}]\url{https://aclanthology.org/D17-1114}
  \end{APACrefURL}
\newblock
\begin{APACrefDOI} \doi{10.18653/v1/D17-1114} \end{APACrefDOI}
\PrintBackRefs{\CurrentBib}

\bibitem [\protect \citeauthoryear {%
M.~Li%
, Zhang%
, Ji%
\BCBL {}\ \BBA {} Radke%
}{%
M.~Li%
\ \protect \BOthers {.}}{%
{\protect \APACyear {2019}}%
}]{%
li_keep_2019}
\APACinsertmetastar {%
li_keep_2019}%
\begin{APACrefauthors}%
Li, M.%
, Zhang, L.%
, Ji, H.%
\BCBL {}\ \BBA {} Radke, R\BPBI J.%
\end{APACrefauthors}%
\unskip\
\newblock
\APACrefYearMonthDay{2019}{{\APACmonth{07}}}{}.
\newblock
{\BBOQ}\APACrefatitle {Keep {Meeting} {Summaries} on {Topic}: {Abstractive}
  {Multi}-{Modal} {Meeting} {Summarization}} {Keep {Meeting} {Summaries} on
  {Topic}: {Abstractive} {Multi}-{Modal} {Meeting} {Summarization}}.{\BBCQ}
\newblock
\BIn{} \APACrefbtitle {Proceedings of the 57th {Annual} {Meeting} of the
  {Association} for {Computational} {Linguistics}} {Proceedings of the 57th
  {Annual} {Meeting} of the {Association} for {Computational} {Linguistics}}\
  (\BPGS\ 2190--2196).
\newblock
\APACaddressPublisher{Florence, Italy}{Association for Computational
  Linguistics}.
\newblock
\begin{APACrefURL} [{2023-05-01}]\url{https://aclanthology.org/P19-1210}
  \end{APACrefURL}
\newblock
\begin{APACrefDOI} \doi{10.18653/v1/P19-1210} \end{APACrefDOI}
\PrintBackRefs{\CurrentBib}

\bibitem [\protect \citeauthoryear {%
W.~Li%
\ \protect \BOthers {.}}{%
W.~Li%
\ \protect \BOthers {.}}{%
{\protect \APACyear {2020}}%
}]{%
li_leveraging_2020}
\APACinsertmetastar {%
li_leveraging_2020}%
\begin{APACrefauthors}%
Li, W.%
, Xiao, X.%
, Liu, J.%
, Wu, H.%
, Wang, H.%
\BCBL {}\ \BBA {} Du, J.%
\end{APACrefauthors}%
\unskip\
\newblock
\APACrefYearMonthDay{2020}{{\APACmonth{07}}}{}.
\newblock
{\BBOQ}\APACrefatitle {Leveraging {Graph} to {Improve} {Abstractive}
  {Multi}-{Document} {Summarization}} {Leveraging {Graph} to {Improve}
  {Abstractive} {Multi}-{Document} {Summarization}}.{\BBCQ}
\newblock
\BIn{} \APACrefbtitle {Proceedings of the 58th {Annual} {Meeting} of the
  {Association} for {Computational} {Linguistics}} {Proceedings of the 58th
  {Annual} {Meeting} of the {Association} for {Computational} {Linguistics}}\
  (\BPGS\ 6232--6243).
\newblock
\APACaddressPublisher{Online}{Association for Computational Linguistics}.
\newblock
\begin{APACrefURL}
  [{2023-05-06}]\url{https://aclanthology.org/2020.acl-main.555}
  \end{APACrefURL}
\newblock
\begin{APACrefDOI} \doi{10.18653/v1/2020.acl-main.555} \end{APACrefDOI}
\PrintBackRefs{\CurrentBib}

\bibitem [\protect \citeauthoryear {%
Lin%
}{%
Lin%
}{%
{\protect \APACyear {2004}}%
}]{%
lin_rouge_2004}
\APACinsertmetastar {%
lin_rouge_2004}%
\begin{APACrefauthors}%
Lin, C\BHBI Y.%
\end{APACrefauthors}%
\unskip\
\newblock
\APACrefYearMonthDay{2004}{{\APACmonth{07}}}{}.
\newblock
{\BBOQ}\APACrefatitle {{ROUGE}: {A} {Package} for {Automatic} {Evaluation} of
  {Summaries}} {{ROUGE}: {A} {Package} for {Automatic} {Evaluation} of
  {Summaries}}.{\BBCQ}
\newblock
\BIn{} \APACrefbtitle {Text {Summarization} {Branches} {Out}} {Text
  {Summarization} {Branches} {Out}}\ (\BPGS\ 74--81).
\newblock
\APACaddressPublisher{Barcelona, Spain}{Association for Computational
  Linguistics}.
\newblock
\begin{APACrefURL} [{2023-07-27}]\url{https://aclanthology.org/W04-1013}
  \end{APACrefURL}
\PrintBackRefs{\CurrentBib}

\bibitem [\protect \citeauthoryear {%
P\BPBI J.~Liu%
\ \protect \BOthers {.}}{%
P\BPBI J.~Liu%
\ \protect \BOthers {.}}{%
{\protect \APACyear {2018}}%
}]{%
liu_generating_2018}
\APACinsertmetastar {%
liu_generating_2018}%
\begin{APACrefauthors}%
Liu, P\BPBI J.%
, Saleh, M.%
, Pot, E.%
, Goodrich, B.%
, Sepassi, R.%
, Kaiser, L.%
\BCBL {}\ \BBA {} Shazeer, N.%
\end{APACrefauthors}%
\unskip\
\newblock
\APACrefYearMonthDay{2018}{{\APACmonth{02}}}{}.
\newblock
{\BBOQ}\APACrefatitle {Generating {Wikipedia} by {Summarizing} {Long}
  {Sequences}} {Generating {Wikipedia} by {Summarizing} {Long}
  {Sequences}}.{\BBCQ}
\newblock
\BIn{} \APACrefbtitle {6th {International} {Conference} on {Learning}
  {Representations}, {ICLR} 2018, {Vancouver}, {BC}, {Canada}, {April} 30 -
  {May} 3, 2018, {Conference} {Track} {Proceedings}.} {6th {International}
  {Conference} on {Learning} {Representations}, {ICLR} 2018, {Vancouver}, {BC},
  {Canada}, {April} 30 - {May} 3, 2018, {Conference} {Track} {Proceedings}.}
\newblock
\begin{APACrefURL}
  [{2023-07-22}]\url{https://openreview.net/forum?id=Hyg0vbWC-}
  \end{APACrefURL}
\PrintBackRefs{\CurrentBib}

\bibitem [\protect \citeauthoryear {%
Y.~Liu%
\ \protect \BOthers {.}}{%
Y.~Liu%
\ \protect \BOthers {.}}{%
{\protect \APACyear {2023}}%
}]{%
liu_revisiting_2023}
\APACinsertmetastar {%
liu_revisiting_2023}%
\begin{APACrefauthors}%
Liu, Y.%
, Fabbri, A.%
, Liu, P.%
, Zhao, Y.%
, Nan, L.%
, Han, R.%
\BDBL {}Radev, D.%
\end{APACrefauthors}%
\unskip\
\newblock
\APACrefYearMonthDay{2023}{{\APACmonth{07}}}{}.
\newblock
{\BBOQ}\APACrefatitle {Revisiting the {Gold} {Standard}: {Grounding}
  {Summarization} {Evaluation} with {Robust} {Human} {Evaluation}} {Revisiting
  the {Gold} {Standard}: {Grounding} {Summarization} {Evaluation} with {Robust}
  {Human} {Evaluation}}.{\BBCQ}
\newblock
\BIn{} \APACrefbtitle {Proceedings of the 61st {Annual} {Meeting} of the
  {Association} for {Computational} {Linguistics} ({Volume} 1: {Long}
  {Papers})} {Proceedings of the 61st {Annual} {Meeting} of the {Association}
  for {Computational} {Linguistics} ({Volume} 1: {Long} {Papers})}\ (\BPGS\
  4140--4170).
\newblock
\APACaddressPublisher{Toronto, Canada}{Association for Computational
  Linguistics}.
\newblock
\begin{APACrefURL}
  [{2023-07-20}]\url{https://aclanthology.org/2023.acl-long.228}
  \end{APACrefURL}
\PrintBackRefs{\CurrentBib}

\bibitem [\protect \citeauthoryear {%
Y.~Liu%
, Jia%
\BCBL {}\ \BBA {} Zhu%
}{%
Y.~Liu%
, Jia%
\BCBL {}\ \BBA {} Zhu%
}{%
{\protect \APACyear {2022}}%
}]{%
liu_length_2022}
\APACinsertmetastar {%
liu_length_2022}%
\begin{APACrefauthors}%
Liu, Y.%
, Jia, Q.%
\BCBL {}\ \BBA {} Zhu, K.%
\end{APACrefauthors}%
\unskip\
\newblock
\APACrefYearMonthDay{2022}{{\APACmonth{05}}}{}.
\newblock
{\BBOQ}\APACrefatitle {Length {Control} in {Abstractive} {Summarization} by
  {Pretraining} {Information} {Selection}} {Length {Control} in {Abstractive}
  {Summarization} by {Pretraining} {Information} {Selection}}.{\BBCQ}
\newblock
\BIn{} \APACrefbtitle {Proceedings of the 60th {Annual} {Meeting} of the
  {Association} for {Computational} {Linguistics} ({Volume} 1: {Long}
  {Papers})} {Proceedings of the 60th {Annual} {Meeting} of the {Association}
  for {Computational} {Linguistics} ({Volume} 1: {Long} {Papers})}\ (\BPGS\
  6885--6895).
\newblock
\APACaddressPublisher{Dublin, Ireland}{Association for Computational
  Linguistics}.
\newblock
\begin{APACrefURL}
  [{2023-03-07}]\url{https://aclanthology.org/2022.acl-long.474}
  \end{APACrefURL}
\newblock
\begin{APACrefDOI} \doi{10.18653/v1/2022.acl-long.474} \end{APACrefDOI}
\PrintBackRefs{\CurrentBib}

\bibitem [\protect \citeauthoryear {%
Y.~Liu%
\ \BBA {} Lapata%
}{%
Y.~Liu%
\ \BBA {} Lapata%
}{%
{\protect \APACyear {2019}}%
{\protect \APACexlab {{\protect \BCnt {1}}}}}]{%
liu_hierarchical_2019}
\APACinsertmetastar {%
liu_hierarchical_2019}%
\begin{APACrefauthors}%
Liu, Y.%
\BCBT {}\ \BBA {} Lapata, M.%
\end{APACrefauthors}%
\unskip\
\newblock
\APACrefYearMonthDay{2019{\protect \BCnt {1}}}{{\APACmonth{07}}}{}.
\newblock
{\BBOQ}\APACrefatitle {Hierarchical {Transformers} for {Multi}-{Document}
  {Summarization}} {Hierarchical {Transformers} for {Multi}-{Document}
  {Summarization}}.{\BBCQ}
\newblock
\BIn{} \APACrefbtitle {Proceedings of the 57th {Annual} {Meeting} of the
  {Association} for {Computational} {Linguistics}} {Proceedings of the 57th
  {Annual} {Meeting} of the {Association} for {Computational} {Linguistics}}\
  (\BPGS\ 5070--5081).
\newblock
\APACaddressPublisher{Florence, Italy}{Association for Computational
  Linguistics}.
\newblock
\begin{APACrefURL} [{2023-05-06}]\url{https://aclanthology.org/P19-1500}
  \end{APACrefURL}
\newblock
\begin{APACrefDOI} \doi{10.18653/v1/P19-1500} \end{APACrefDOI}
\PrintBackRefs{\CurrentBib}

\bibitem [\protect \citeauthoryear {%
Y.~Liu%
\ \BBA {} Lapata%
}{%
Y.~Liu%
\ \BBA {} Lapata%
}{%
{\protect \APACyear {2019}}%
{\protect \APACexlab {{\protect \BCnt {2}}}}}]{%
liu_text_2019}
\APACinsertmetastar {%
liu_text_2019}%
\begin{APACrefauthors}%
Liu, Y.%
\BCBT {}\ \BBA {} Lapata, M.%
\end{APACrefauthors}%
\unskip\
\newblock
\APACrefYearMonthDay{2019{\protect \BCnt {2}}}{{\APACmonth{11}}}{}.
\newblock
{\BBOQ}\APACrefatitle {Text {Summarization} with {Pretrained} {Encoders}} {Text
  {Summarization} with {Pretrained} {Encoders}}.{\BBCQ}
\newblock
\BIn{} \APACrefbtitle {Proceedings of the 2019 {Conference} on {Empirical}
  {Methods} in {Natural} {Language} {Processing} and the 9th {International}
  {Joint} {Conference} on {Natural} {Language} {Processing} ({EMNLP}-{IJCNLP})}
  {Proceedings of the 2019 {Conference} on {Empirical} {Methods} in {Natural}
  {Language} {Processing} and the 9th {International} {Joint} {Conference} on
  {Natural} {Language} {Processing} ({EMNLP}-{IJCNLP})}\ (\BPGS\ 3730--3740).
\newblock
\APACaddressPublisher{Hong Kong, China}{Association for Computational
  Linguistics}.
\newblock
\begin{APACrefURL} [{2023-07-20}]\url{https://aclanthology.org/D19-1387}
  \end{APACrefURL}
\newblock
\begin{APACrefDOI} \doi{10.18653/v1/D19-1387} \end{APACrefDOI}
\PrintBackRefs{\CurrentBib}

\bibitem [\protect \citeauthoryear {%
Y.~Liu%
, Liu%
, Radev%
\BCBL {}\ \BBA {} Neubig%
}{%
Y.~Liu%
, Liu%
\BCBL {}\ \protect \BOthers {.}}{%
{\protect \APACyear {2022}}%
}]{%
liu_brio_2022}
\APACinsertmetastar {%
liu_brio_2022}%
\begin{APACrefauthors}%
Liu, Y.%
, Liu, P.%
, Radev, D.%
\BCBL {}\ \BBA {} Neubig, G.%
\end{APACrefauthors}%
\unskip\
\newblock
\APACrefYearMonthDay{2022}{{\APACmonth{05}}}{}.
\newblock
{\BBOQ}\APACrefatitle {{BRIO}: {Bringing} {Order} to {Abstractive}
  {Summarization}} {{BRIO}: {Bringing} {Order} to {Abstractive}
  {Summarization}}.{\BBCQ}
\newblock
\BIn{} \APACrefbtitle {Proceedings of the 60th {Annual} {Meeting} of the
  {Association} for {Computational} {Linguistics} ({Volume} 1: {Long}
  {Papers})} {Proceedings of the 60th {Annual} {Meeting} of the {Association}
  for {Computational} {Linguistics} ({Volume} 1: {Long} {Papers})}\ (\BPGS\
  2890--2903).
\newblock
\APACaddressPublisher{Dublin, Ireland}{Association for Computational
  Linguistics}.
\newblock
\begin{APACrefURL}
  [{2023-07-21}]\url{https://aclanthology.org/2022.acl-long.207}
  \end{APACrefURL}
\newblock
\begin{APACrefDOI} \doi{10.18653/v1/2022.acl-long.207} \end{APACrefDOI}
\PrintBackRefs{\CurrentBib}

\bibitem [\protect \citeauthoryear {%
Y.~Liu%
, Luo%
\BCBL {}\ \BBA {} Zhu%
}{%
Y.~Liu%
\ \protect \BOthers {.}}{%
{\protect \APACyear {2018}}%
}]{%
liu_controlling_2018}
\APACinsertmetastar {%
liu_controlling_2018}%
\begin{APACrefauthors}%
Liu, Y.%
, Luo, Z.%
\BCBL {}\ \BBA {} Zhu, K.%
\end{APACrefauthors}%
\unskip\
\newblock
\APACrefYearMonthDay{2018}{{\APACmonth{10}}}{}.
\newblock
{\BBOQ}\APACrefatitle {Controlling {Length} in {Abstractive} {Summarization}
  {Using} a {Convolutional} {Neural} {Network}} {Controlling {Length} in
  {Abstractive} {Summarization} {Using} a {Convolutional} {Neural}
  {Network}}.{\BBCQ}
\newblock
\BIn{} \APACrefbtitle {Proceedings of the 2018 {Conference} on {Empirical}
  {Methods} in {Natural} {Language} {Processing}} {Proceedings of the 2018
  {Conference} on {Empirical} {Methods} in {Natural} {Language} {Processing}}\
  (\BPGS\ 4110--4119).
\newblock
\APACaddressPublisher{Brussels, Belgium}{Association for Computational
  Linguistics}.
\newblock
\begin{APACrefURL} [{2023-03-21}]\url{https://aclanthology.org/D18-1444}
  \end{APACrefURL}
\newblock
\begin{APACrefDOI} \doi{10.18653/v1/D18-1444} \end{APACrefDOI}
\PrintBackRefs{\CurrentBib}

\bibitem [\protect \citeauthoryear {%
Midha%
}{%
Midha%
}{%
{\protect \APACyear {2023}}%
}]{%
midha_discord_2023}
\APACinsertmetastar {%
midha_discord_2023}%
\begin{APACrefauthors}%
Midha, A.%
\end{APACrefauthors}%
\unskip\
\newblock
\APACrefYearMonthDay{2023}{{\APACmonth{03}}}{}.
\newblock
\APACrefbtitle {Discord is {Your} {Place} for {AI} with {Friends}.} {Discord is
  {Your} {Place} for {AI} with {Friends}.}
\newblock
\begin{APACrefURL}
  [{2023-03-09}]\url{https://discord.com/blog/ai-on-discord-your-place-for-ai-with-friends}
  \end{APACrefURL}
\PrintBackRefs{\CurrentBib}

\bibitem [\protect \citeauthoryear {%
Muennighoff%
\ \protect \BOthers {.}}{%
Muennighoff%
\ \protect \BOthers {.}}{%
{\protect \APACyear {2023}}%
}]{%
muennighoff_crosslingual_2023}
\APACinsertmetastar {%
muennighoff_crosslingual_2023}%
\begin{APACrefauthors}%
Muennighoff, N.%
, Wang, T.%
, Sutawika, L.%
, Roberts, A.%
, Biderman, S.%
, Le~Scao, T.%
\BDBL {}Raffel, C.%
\end{APACrefauthors}%
\unskip\
\newblock
\APACrefYearMonthDay{2023}{{\APACmonth{07}}}{}.
\newblock
{\BBOQ}\APACrefatitle {Crosslingual {Generalization} through {Multitask}
  {Finetuning}} {Crosslingual {Generalization} through {Multitask}
  {Finetuning}}.{\BBCQ}
\newblock
\BIn{} \APACrefbtitle {Proceedings of the 61st {Annual} {Meeting} of the
  {Association} for {Computational} {Linguistics} ({Volume} 1: {Long}
  {Papers})} {Proceedings of the 61st {Annual} {Meeting} of the {Association}
  for {Computational} {Linguistics} ({Volume} 1: {Long} {Papers})}\ (\BPGS\
  15991--16111).
\newblock
\APACaddressPublisher{Toronto, Canada}{Association for Computational
  Linguistics}.
\newblock
\begin{APACrefURL}
  [{2023-07-21}]\url{https://aclanthology.org/2023.acl-long.891}
  \end{APACrefURL}
\PrintBackRefs{\CurrentBib}

\bibitem [\protect \citeauthoryear {%
Nallapati%
, Zhou%
, santos%
, Gulcehre%
\BCBL {}\ \BBA {} Xiang%
}{%
Nallapati%
\ \protect \BOthers {.}}{%
{\protect \APACyear {2016}}%
}]{%
nallapati_abstractive_2016}
\APACinsertmetastar {%
nallapati_abstractive_2016}%
\begin{APACrefauthors}%
Nallapati, R.%
, Zhou, B.%
, santos, C\BPBI N\BPBI d.%
, Gulcehre, C.%
\BCBL {}\ \BBA {} Xiang, B.%
\end{APACrefauthors}%
\unskip\
\newblock
\APACrefYearMonthDay{2016}{{\APACmonth{08}}}{}.
\newblock
\APACrefbtitle {Abstractive {Text} {Summarization} {Using}
  {Sequence}-to-{Sequence} {RNNs} and {Beyond}.} {Abstractive {Text}
  {Summarization} {Using} {Sequence}-to-{Sequence} {RNNs} and {Beyond}.}
\newblock
\APACaddressPublisher{}{arXiv}.
\newblock
\begin{APACrefURL} [{2023-03-01}]\url{http://arxiv.org/abs/1602.06023}
  \end{APACrefURL}
\newblock
\APACrefnote{arXiv:1602.06023 [cs] version: 5}
\PrintBackRefs{\CurrentBib}

\bibitem [\protect \citeauthoryear {%
Ouyang%
\ \protect \BOthers {.}}{%
Ouyang%
\ \protect \BOthers {.}}{%
{\protect \APACyear {2022}}%
}]{%
ouyang_training_2022}
\APACinsertmetastar {%
ouyang_training_2022}%
\begin{APACrefauthors}%
Ouyang, L.%
, Wu, J.%
, Jiang, X.%
, Almeida, D.%
, Wainwright, C.%
, Mishkin, P.%
\BDBL {}Lowe, R.%
\end{APACrefauthors}%
\unskip\
\newblock
\APACrefYearMonthDay{2022}{}{}.
\newblock
{\BBOQ}\APACrefatitle {Training language models to follow instructions with
  human feedback} {Training language models to follow instructions with human
  feedback}.{\BBCQ}
\newblock
\BIn{} S.~Koyejo, S.~Mohamed, A.~Agarwal, D.~Belgrave, K.~Cho\BCBL {}\ \BBA {}
  A.~Oh\ (\BEDS), \APACrefbtitle {Advances in {Neural} {Information}
  {Processing} {Systems}} {Advances in {Neural} {Information} {Processing}
  {Systems}}\ (\BVOL~35, \BPGS\ 27730--27744).
\newblock
\APACaddressPublisher{}{Curran Associates, Inc.}
\newblock
\begin{APACrefURL}
  \url{https://proceedings.neurips.cc/paper_files/paper/2022/file/b1efde53be364a73914f58805a001731-Paper-Conference.pdf}
  \end{APACrefURL}
\PrintBackRefs{\CurrentBib}

\bibitem [\protect \citeauthoryear {%
Pagnoni%
, Balachandran%
\BCBL {}\ \BBA {} Tsvetkov%
}{%
Pagnoni%
\ \protect \BOthers {.}}{%
{\protect \APACyear {2021}}%
}]{%
pagnoni_understanding_2021}
\APACinsertmetastar {%
pagnoni_understanding_2021}%
\begin{APACrefauthors}%
Pagnoni, A.%
, Balachandran, V.%
\BCBL {}\ \BBA {} Tsvetkov, Y.%
\end{APACrefauthors}%
\unskip\
\newblock
\APACrefYearMonthDay{2021}{{\APACmonth{06}}}{}.
\newblock
{\BBOQ}\APACrefatitle {Understanding {Factuality} in {Abstractive}
  {Summarization} with {FRANK}: {A} {Benchmark} for {Factuality} {Metrics}}
  {Understanding {Factuality} in {Abstractive} {Summarization} with {FRANK}:
  {A} {Benchmark} for {Factuality} {Metrics}}.{\BBCQ}
\newblock
\BIn{} \APACrefbtitle {Proceedings of the 2021 {Conference} of the {North}
  {American} {Chapter} of the {Association} for {Computational} {Linguistics}:
  {Human} {Language} {Technologies}} {Proceedings of the 2021 {Conference} of
  the {North} {American} {Chapter} of the {Association} for {Computational}
  {Linguistics}: {Human} {Language} {Technologies}}\ (\BPGS\ 4812--4829).
\newblock
\APACaddressPublisher{Online}{Association for Computational Linguistics}.
\newblock
\begin{APACrefURL}
  [{2023-07-20}]\url{https://aclanthology.org/2021.naacl-main.383}
  \end{APACrefURL}
\newblock
\begin{APACrefDOI} \doi{10.18653/v1/2021.naacl-main.383} \end{APACrefDOI}
\PrintBackRefs{\CurrentBib}

\bibitem [\protect \citeauthoryear {%
Parthasarathy%
}{%
Parthasarathy%
}{%
{\protect \APACyear {2023}}%
}]{%
parthasarathy_zooms_2023}
\APACinsertmetastar {%
parthasarathy_zooms_2023}%
\begin{APACrefauthors}%
Parthasarathy, V.%
\end{APACrefauthors}%
\unskip\
\newblock
\APACrefYearMonthDay{2023}{{\APACmonth{02}}}{}.
\newblock
\APACrefbtitle {Zoom’s {AI} innovations empower people.} {Zoom’s {AI}
  innovations empower people.}
\newblock
\begin{APACrefURL}
  [{2023-03-16}]\url{https://blog.zoom.us/ai-driven-innovations/}
  \end{APACrefURL}
\PrintBackRefs{\CurrentBib}

\bibitem [\protect \citeauthoryear {%
Pasunuru%
, Liu%
, Bansal%
, Ravi%
\BCBL {}\ \BBA {} Dreyer%
}{%
Pasunuru%
\ \protect \BOthers {.}}{%
{\protect \APACyear {2021}}%
}]{%
pasunuru_efficiently_2021}
\APACinsertmetastar {%
pasunuru_efficiently_2021}%
\begin{APACrefauthors}%
Pasunuru, R.%
, Liu, M.%
, Bansal, M.%
, Ravi, S.%
\BCBL {}\ \BBA {} Dreyer, M.%
\end{APACrefauthors}%
\unskip\
\newblock
\APACrefYearMonthDay{2021}{{\APACmonth{06}}}{}.
\newblock
{\BBOQ}\APACrefatitle {Efficiently {Summarizing} {Text} and {Graph} {Encodings}
  of {Multi}-{Document} {Clusters}} {Efficiently {Summarizing} {Text} and
  {Graph} {Encodings} of {Multi}-{Document} {Clusters}}.{\BBCQ}
\newblock
\BIn{} \APACrefbtitle {Proceedings of the 2021 {Conference} of the {North}
  {American} {Chapter} of the {Association} for {Computational} {Linguistics}:
  {Human} {Language} {Technologies}} {Proceedings of the 2021 {Conference} of
  the {North} {American} {Chapter} of the {Association} for {Computational}
  {Linguistics}: {Human} {Language} {Technologies}}\ (\BPGS\ 4768--4779).
\newblock
\APACaddressPublisher{Online}{Association for Computational Linguistics}.
\newblock
\begin{APACrefURL}
  [{2023-05-06}]\url{https://aclanthology.org/2021.naacl-main.380}
  \end{APACrefURL}
\newblock
\begin{APACrefDOI} \doi{10.18653/v1/2021.naacl-main.380} \end{APACrefDOI}
\PrintBackRefs{\CurrentBib}

\bibitem [\protect \citeauthoryear {%
Qin%
\ \protect \BOthers {.}}{%
Qin%
\ \protect \BOthers {.}}{%
{\protect \APACyear {2023}}%
}]{%
qin_is_2023}
\APACinsertmetastar {%
qin_is_2023}%
\begin{APACrefauthors}%
Qin, C.%
, Zhang, A.%
, Zhang, Z.%
, Chen, J.%
, Yasunaga, M.%
\BCBL {}\ \BBA {} Yang, D.%
\end{APACrefauthors}%
\unskip\
\newblock
\APACrefYearMonthDay{2023}{{\APACmonth{02}}}{}.
\newblock
\APACrefbtitle {Is {ChatGPT} a {General}-{Purpose} {Natural} {Language}
  {Processing} {Task} {Solver}?} {Is {ChatGPT} a {General}-{Purpose} {Natural}
  {Language} {Processing} {Task} {Solver}?}
\newblock
\APACaddressPublisher{}{arXiv}.
\newblock
\begin{APACrefURL} [{2023-03-20}]\url{http://arxiv.org/abs/2302.06476}
  \end{APACrefURL}
\newblock
\APACrefnote{arXiv:2302.06476 [cs]}
\PrintBackRefs{\CurrentBib}

\bibitem [\protect \citeauthoryear {%
Rae%
\ \protect \BOthers {.}}{%
Rae%
\ \protect \BOthers {.}}{%
{\protect \APACyear {2022}}%
}]{%
rae_scaling_2022}
\APACinsertmetastar {%
rae_scaling_2022}%
\begin{APACrefauthors}%
Rae, J\BPBI W.%
, Borgeaud, S.%
, Cai, T.%
, Millican, K.%
, Hoffmann, J.%
, Song, F.%
\BDBL {}Irving, G.%
\end{APACrefauthors}%
\unskip\
\newblock
\APACrefYearMonthDay{2022}{{\APACmonth{01}}}{}.
\newblock
\APACrefbtitle {Scaling {Language} {Models}: {Methods}, {Analysis} \&
  {Insights} from {Training} {Gopher}.} {Scaling {Language} {Models}:
  {Methods}, {Analysis} \& {Insights} from {Training} {Gopher}.}
\newblock
\APACaddressPublisher{}{arXiv}.
\newblock
\begin{APACrefURL} [{2023-02-28}]\url{http://arxiv.org/abs/2112.11446}
  \end{APACrefURL}
\newblock
\APACrefnote{arXiv:2112.11446 [cs]}
\PrintBackRefs{\CurrentBib}

\bibitem [\protect \citeauthoryear {%
Raffel%
\ \protect \BOthers {.}}{%
Raffel%
\ \protect \BOthers {.}}{%
{\protect \APACyear {2020}}%
}]{%
raffel_exploring_2020}
\APACinsertmetastar {%
raffel_exploring_2020}%
\begin{APACrefauthors}%
Raffel, C.%
, Shazeer, N.%
, Roberts, A.%
, Lee, K.%
, Narang, S.%
, Matena, M.%
\BDBL {}Liu, P\BPBI J.%
\end{APACrefauthors}%
\unskip\
\newblock
\APACrefYearMonthDay{2020}{}{}.
\newblock
{\BBOQ}\APACrefatitle {Exploring the {Limits} of {Transfer} {Learning} with a
  {Unified} {Text}-to-{Text} {Transformer}} {Exploring the {Limits} of
  {Transfer} {Learning} with a {Unified} {Text}-to-{Text}
  {Transformer}}.{\BBCQ}
\newblock
\APACjournalVolNumPages{Journal of Machine Learning Research}{21}{140}{1--67}.
\newblock
\begin{APACrefURL} [{2023-07-20}]\url{http://jmlr.org/papers/v21/20-074.html}
  \end{APACrefURL}
\PrintBackRefs{\CurrentBib}

\bibitem [\protect \citeauthoryear {%
Reed%
\ \protect \BOthers {.}}{%
Reed%
\ \protect \BOthers {.}}{%
{\protect \APACyear {2022}}%
}]{%
reed_generalist_2022}
\APACinsertmetastar {%
reed_generalist_2022}%
\begin{APACrefauthors}%
Reed, S.%
, Zolna, K.%
, Parisotto, E.%
, Colmenarejo, S\BPBI G.%
, Novikov, A.%
, Barth-maron, G.%
\BDBL {}Freitas, N\BPBI d.%
\end{APACrefauthors}%
\unskip\
\newblock
\APACrefYearMonthDay{2022}{{\APACmonth{08}}}{}.
\newblock
{\BBOQ}\APACrefatitle {A {Generalist} {Agent}} {A {Generalist} {Agent}}.{\BBCQ}
\newblock
\APACjournalVolNumPages{Transactions on Machine Learning Research}{}{}{}.
\newblock
\begin{APACrefURL}
  [{2023-07-22}]\url{https://openreview.net/forum?id=1ikK0kHjvj}
  \end{APACrefURL}
\PrintBackRefs{\CurrentBib}

\bibitem [\protect \citeauthoryear {%
Saito%
\ \protect \BOthers {.}}{%
Saito%
\ \protect \BOthers {.}}{%
{\protect \APACyear {2020}}%
}]{%
saito_length-controllable_2020}
\APACinsertmetastar {%
saito_length-controllable_2020}%
\begin{APACrefauthors}%
Saito, I.%
, Nishida, K.%
, Nishida, K.%
, Otsuka, A.%
, Asano, H.%
, Tomita, J.%
\BDBL {}Matsumoto, Y.%
\end{APACrefauthors}%
\unskip\
\newblock
\APACrefYearMonthDay{2020}{{\APACmonth{01}}}{}.
\newblock
\APACrefbtitle {Length-controllable {Abstractive} {Summarization} by {Guiding}
  with {Summary} {Prototype}.} {Length-controllable {Abstractive}
  {Summarization} by {Guiding} with {Summary} {Prototype}.}
\newblock
\APACaddressPublisher{}{arXiv}.
\newblock
\begin{APACrefURL} [{2023-03-21}]\url{http://arxiv.org/abs/2001.07331}
  \end{APACrefURL}
\newblock
\APACrefnote{arXiv:2001.07331 [cs]}
\newblock
\begin{APACrefDOI} \doi{10.48550/arXiv.2001.07331} \end{APACrefDOI}
\PrintBackRefs{\CurrentBib}

\bibitem [\protect \citeauthoryear {%
Saleh%
\ \BBA {} Kannan%
}{%
Saleh%
\ \BBA {} Kannan%
}{%
{\protect \APACyear {2022}}%
}]{%
saleh_auto-generated_2022}
\APACinsertmetastar {%
saleh_auto-generated_2022}%
\begin{APACrefauthors}%
Saleh, M.%
\BCBT {}\ \BBA {} Kannan, A.%
\end{APACrefauthors}%
\unskip\
\newblock
\APACrefYearMonthDay{2022}{{\APACmonth{03}}}{}.
\newblock
\APACrefbtitle {Auto-generated {Summaries} in {Google} {Docs}.} {Auto-generated
  {Summaries} in {Google} {Docs}.}
\newblock
\begin{APACrefURL}
  [{2023-02-28}]\url{https://ai.googleblog.com/2022/03/auto-generated-summaries-in-google-docs.html}
  \end{APACrefURL}
\PrintBackRefs{\CurrentBib}

\bibitem [\protect \citeauthoryear {%
Saleh%
\ \BBA {} Wang%
}{%
Saleh%
\ \BBA {} Wang%
}{%
{\protect \APACyear {2022}}%
}]{%
saleh_conversation_2022}
\APACinsertmetastar {%
saleh_conversation_2022}%
\begin{APACrefauthors}%
Saleh, M.%
\BCBT {}\ \BBA {} Wang, Y.%
\end{APACrefauthors}%
\unskip\
\newblock
\APACrefYearMonthDay{2022}{{\APACmonth{11}}}{}.
\newblock
\APACrefbtitle {Conversation {Summaries} in {Google} {Chat}.} {Conversation
  {Summaries} in {Google} {Chat}.}
\newblock
\begin{APACrefURL}
  [{2023-02-28}]\url{https://ai.googleblog.com/2022/11/conversation-summaries-in-google-chat.html}
  \end{APACrefURL}
\PrintBackRefs{\CurrentBib}

\bibitem [\protect \citeauthoryear {%
Sanh%
\ \protect \BOthers {.}}{%
Sanh%
\ \protect \BOthers {.}}{%
{\protect \APACyear {2022}}%
}]{%
sanh_multitask_2022}
\APACinsertmetastar {%
sanh_multitask_2022}%
\begin{APACrefauthors}%
Sanh, V.%
, Webson, A.%
, Raffel, C.%
, Bach, S\BPBI H.%
, Sutawika, L.%
, Alyafeai, Z.%
\BDBL {}Rush, A\BPBI M.%
\end{APACrefauthors}%
\unskip\
\newblock
\APACrefYearMonthDay{2022}{{\APACmonth{03}}}{}.
\newblock
\APACrefbtitle {Multitask {Prompted} {Training} {Enables} {Zero}-{Shot} {Task}
  {Generalization}.} {Multitask {Prompted} {Training} {Enables} {Zero}-{Shot}
  {Task} {Generalization}.}
\newblock
\APACaddressPublisher{}{arXiv}.
\newblock
\begin{APACrefURL} [{2023-03-01}]\url{http://arxiv.org/abs/2110.08207}
  \end{APACrefURL}
\newblock
\APACrefnote{arXiv:2110.08207 [cs]}
\PrintBackRefs{\CurrentBib}

\bibitem [\protect \citeauthoryear {%
See%
, Liu%
\BCBL {}\ \BBA {} Manning%
}{%
See%
\ \protect \BOthers {.}}{%
{\protect \APACyear {2017}}%
}]{%
see_get_2017}
\APACinsertmetastar {%
see_get_2017}%
\begin{APACrefauthors}%
See, A.%
, Liu, P\BPBI J.%
\BCBL {}\ \BBA {} Manning, C\BPBI D.%
\end{APACrefauthors}%
\unskip\
\newblock
\APACrefYearMonthDay{2017}{{\APACmonth{07}}}{}.
\newblock
{\BBOQ}\APACrefatitle {Get {To} {The} {Point}: {Summarization} with
  {Pointer}-{Generator} {Networks}} {Get {To} {The} {Point}: {Summarization}
  with {Pointer}-{Generator} {Networks}}.{\BBCQ}
\newblock
\BIn{} \APACrefbtitle {Proceedings of the 55th {Annual} {Meeting} of the
  {Association} for {Computational} {Linguistics} ({Volume} 1: {Long}
  {Papers})} {Proceedings of the 55th {Annual} {Meeting} of the {Association}
  for {Computational} {Linguistics} ({Volume} 1: {Long} {Papers})}\ (\BPGS\
  1073--1083).
\newblock
\APACaddressPublisher{Vancouver, Canada}{Association for Computational
  Linguistics}.
\newblock
\begin{APACrefURL} [{2023-07-20}]\url{https://aclanthology.org/P17-1099}
  \end{APACrefURL}
\newblock
\begin{APACrefDOI} \doi{10.18653/v1/P17-1099} \end{APACrefDOI}
\PrintBackRefs{\CurrentBib}

\bibitem [\protect \citeauthoryear {%
Sharma%
, Li%
\BCBL {}\ \BBA {} Wang%
}{%
Sharma%
\ \protect \BOthers {.}}{%
{\protect \APACyear {2019}}%
}]{%
sharma_bigpatent_2019}
\APACinsertmetastar {%
sharma_bigpatent_2019}%
\begin{APACrefauthors}%
Sharma, E.%
, Li, C.%
\BCBL {}\ \BBA {} Wang, L.%
\end{APACrefauthors}%
\unskip\
\newblock
\APACrefYearMonthDay{2019}{{\APACmonth{07}}}{}.
\newblock
{\BBOQ}\APACrefatitle {{BIGPATENT}: {A} {Large}-{Scale} {Dataset} for
  {Abstractive} and {Coherent} {Summarization}} {{BIGPATENT}: {A}
  {Large}-{Scale} {Dataset} for {Abstractive} and {Coherent}
  {Summarization}}.{\BBCQ}
\newblock
\BIn{} \APACrefbtitle {Proceedings of the 57th {Annual} {Meeting} of the
  {Association} for {Computational} {Linguistics}} {Proceedings of the 57th
  {Annual} {Meeting} of the {Association} for {Computational} {Linguistics}}\
  (\BPGS\ 2204--2213).
\newblock
\APACaddressPublisher{Florence, Italy}{Association for Computational
  Linguistics}.
\newblock
\begin{APACrefURL} [{2023-03-19}]\url{https://aclanthology.org/P19-1212}
  \end{APACrefURL}
\newblock
\begin{APACrefDOI} \doi{10.18653/v1/P19-1212} \end{APACrefDOI}
\PrintBackRefs{\CurrentBib}

\bibitem [\protect \citeauthoryear {%
Szyndzielorz%
}{%
Szyndzielorz%
}{%
{\protect \APACyear {2023}}%
}]{%
szyndzielorz_opera_2023}
\APACinsertmetastar {%
szyndzielorz_opera_2023}%
\begin{APACrefauthors}%
Szyndzielorz, J.%
\end{APACrefauthors}%
\unskip\
\newblock
\APACrefYearMonthDay{2023}{{\APACmonth{02}}}{}.
\newblock
\APACrefbtitle {Opera enters the generative {AI} space with new features in
  browsers and content apps.} {Opera enters the generative {AI} space with new
  features in browsers and content apps.}
\newblock
\begin{APACrefURL}
  [{2023-03-01}]\url{https://blogs.opera.com/news/2023/02/opera-aigc-integration/}
  \end{APACrefURL}
\PrintBackRefs{\CurrentBib}

\bibitem [\protect \citeauthoryear {%
Taori%
\ \protect \BOthers {.}}{%
Taori%
\ \protect \BOthers {.}}{%
{\protect \APACyear {2023}}%
}]{%
taori_alpaca_2023}
\APACinsertmetastar {%
taori_alpaca_2023}%
\begin{APACrefauthors}%
Taori, R.%
, Gulrajani, I.%
, Zhang, T.%
, Dubois, Y.%
, Guestrin, C.%
, Liang, P.%
\BCBL {}\ \BBA {} Hashimoto, T\BPBI B.%
\end{APACrefauthors}%
\unskip\
\newblock
\APACrefYearMonthDay{2023}{{\APACmonth{03}}}{}.
\newblock
\APACrefbtitle {Alpaca: {A} {Strong}, {Replicable} {Instruction}-{Following}
  {Model}.} {Alpaca: {A} {Strong}, {Replicable} {Instruction}-{Following}
  {Model}.}
\newblock
\begin{APACrefURL}
  [{2023-05-07}]\url{https://crfm.stanford.edu/2023/03/13/alpaca.html}
  \end{APACrefURL}
\PrintBackRefs{\CurrentBib}

\bibitem [\protect \citeauthoryear {%
Tay%
\ \protect \BOthers {.}}{%
Tay%
\ \protect \BOthers {.}}{%
{\protect \APACyear {2022}}%
}]{%
tay_ul2_2022}
\APACinsertmetastar {%
tay_ul2_2022}%
\begin{APACrefauthors}%
Tay, Y.%
, Dehghani, M.%
, Tran, V\BPBI Q.%
, Garcia, X.%
, Wei, J.%
, Wang, X.%
\BDBL {}Metzler, D.%
\end{APACrefauthors}%
\unskip\
\newblock
\APACrefYearMonthDay{2022}{{\APACmonth{09}}}{}.
\newblock
{\BBOQ}\APACrefatitle {{UL2}: {Unifying} {Language} {Learning} {Paradigms}}
  {{UL2}: {Unifying} {Language} {Learning} {Paradigms}}.{\BBCQ}
\newblock
\BIn{} \APACrefbtitle {The {Eleventh} {International} {Conference} on
  {Learning} {Representations}, {ICLR} 2023, {Kigali}, {Rwanda}, {May} 1-5,
  2023.} {The {Eleventh} {International} {Conference} on {Learning}
  {Representations}, {ICLR} 2023, {Kigali}, {Rwanda}, {May} 1-5, 2023.}
\newblock
\begin{APACrefURL}
  [{2023-07-22}]\url{https://openreview.net/forum?id=6ruVLB727MC}
  \end{APACrefURL}
\PrintBackRefs{\CurrentBib}

\bibitem [\protect \citeauthoryear {%
Taylor%
\ \protect \BOthers {.}}{%
Taylor%
\ \protect \BOthers {.}}{%
{\protect \APACyear {2022}}%
}]{%
taylor_galactica_2022}
\APACinsertmetastar {%
taylor_galactica_2022}%
\begin{APACrefauthors}%
Taylor, R.%
, Kardas, M.%
, Cucurull, G.%
, Scialom, T.%
, Hartshorn, A.%
, Saravia, E.%
\BDBL {}Stojnic, R.%
\end{APACrefauthors}%
\unskip\
\newblock
\APACrefYearMonthDay{2022}{{\APACmonth{11}}}{}.
\newblock
\APACrefbtitle {Galactica: {A} {Large} {Language} {Model} for {Science}.}
  {Galactica: {A} {Large} {Language} {Model} for {Science}.}
\newblock
\APACaddressPublisher{}{arXiv}.
\newblock
\begin{APACrefURL} [{2023-02-28}]\url{http://arxiv.org/abs/2211.09085}
  \end{APACrefURL}
\newblock
\APACrefnote{arXiv:2211.09085 [cs, stat]}
\newblock
\begin{APACrefDOI} \doi{10.48550/arXiv.2211.09085} \end{APACrefDOI}
\PrintBackRefs{\CurrentBib}

\bibitem [\protect \citeauthoryear {%
{The Vicuna Team}%
}{%
{The Vicuna Team}%
}{%
{\protect \APACyear {2023}}%
}]{%
the_vicuna_team_vicuna_2023}
\APACinsertmetastar {%
the_vicuna_team_vicuna_2023}%
\begin{APACrefauthors}%
{The Vicuna Team}.%
\end{APACrefauthors}%
\unskip\
\newblock
\APACrefYearMonthDay{2023}{{\APACmonth{03}}}{}.
\newblock
\APACrefbtitle {Vicuna: {An} {Open}-{Source} {Chatbot} {Impressing} {GPT}-4
  with 90\%* {ChatGPT} {Quality} {\textbar} {LMSYS} {Org}.} {Vicuna: {An}
  {Open}-{Source} {Chatbot} {Impressing} {GPT}-4 with 90\%* {ChatGPT} {Quality}
  {\textbar} {LMSYS} {Org}.}
\newblock
\begin{APACrefURL} [{2023-05-07}]\url{https://lmsys.org/blog/2023-03-30-vicuna}
  \end{APACrefURL}
\PrintBackRefs{\CurrentBib}

\bibitem [\protect \citeauthoryear {%
Toews%
}{%
Toews%
}{%
{\protect \APACyear {2022}}%
}]{%
toews_wave_2022}
\APACinsertmetastar {%
toews_wave_2022}%
\begin{APACrefauthors}%
Toews, R.%
\end{APACrefauthors}%
\unskip\
\newblock
\APACrefYearMonthDay{2022}{{\APACmonth{05}}}{}.
\newblock
\APACrefbtitle {A {Wave} {Of} {Billion}-{Dollar} {Language} {AI} {Startups}
  {Is} {Coming}.} {A {Wave} {Of} {Billion}-{Dollar} {Language} {AI} {Startups}
  {Is} {Coming}.}
\newblock
\begin{APACrefURL}
  [{2023-03-08}]\url{https://www.forbes.com/sites/robtoews/2022/03/27/a-wave-of-billion-dollar-language-ai-startups-is-coming/}
  \end{APACrefURL}
\newblock
\APACrefnote{Section: AI}
\PrintBackRefs{\CurrentBib}

\bibitem [\protect \citeauthoryear {%
Touvron%
\ \protect \BOthers {.}}{%
Touvron%
\ \protect \BOthers {.}}{%
{\protect \APACyear {2023}}%
}]{%
touvron_llama_2023}
\APACinsertmetastar {%
touvron_llama_2023}%
\begin{APACrefauthors}%
Touvron, H.%
, Lavril, T.%
, Izacard, G.%
, Martinet, X.%
, Lachaux, M\BHBI A.%
, Lacroix, T.%
\BDBL {}Lample, G.%
\end{APACrefauthors}%
\unskip\
\newblock
\APACrefYearMonthDay{2023}{}{}.
\newblock
{\BBOQ}\APACrefatitle {{LLaMA}: {Open} and {Efficient} {Foundation} {Language}
  {Models}} {{LLaMA}: {Open} and {Efficient} {Foundation} {Language}
  {Models}}.{\BBCQ}
\newblock

\PrintBackRefs{\CurrentBib}

\bibitem [\protect \citeauthoryear {%
A.~Wang%
, Pang%
, Chen%
, Phang%
\BCBL {}\ \BBA {} Bowman%
}{%
A.~Wang%
\ \protect \BOthers {.}}{%
{\protect \APACyear {2022}}%
}]{%
wang_squality_2022}
\APACinsertmetastar {%
wang_squality_2022}%
\begin{APACrefauthors}%
Wang, A.%
, Pang, R\BPBI Y.%
, Chen, A.%
, Phang, J.%
\BCBL {}\ \BBA {} Bowman, S\BPBI R.%
\end{APACrefauthors}%
\unskip\
\newblock
\APACrefYearMonthDay{2022}{{\APACmonth{12}}}{}.
\newblock
{\BBOQ}\APACrefatitle {{SQuALITY}: {Building} a {Long}-{Document}
  {Summarization} {Dataset} the {Hard} {Way}} {{SQuALITY}: {Building} a
  {Long}-{Document} {Summarization} {Dataset} the {Hard} {Way}}.{\BBCQ}
\newblock
\BIn{} \APACrefbtitle {Proceedings of the 2022 {Conference} on {Empirical}
  {Methods} in {Natural} {Language} {Processing}} {Proceedings of the 2022
  {Conference} on {Empirical} {Methods} in {Natural} {Language} {Processing}}\
  (\BPGS\ 1139--1156).
\newblock
\APACaddressPublisher{Abu Dhabi, United Arab Emirates}{Association for
  Computational Linguistics}.
\newblock
\begin{APACrefURL}
  [{2023-07-20}]\url{https://aclanthology.org/2022.emnlp-main.75}
  \end{APACrefURL}
\PrintBackRefs{\CurrentBib}

\bibitem [\protect \citeauthoryear {%
A.~Wang%
\ \protect \BOthers {.}}{%
A.~Wang%
\ \protect \BOthers {.}}{%
{\protect \APACyear {2019}}%
}]{%
wang_superglue_2019}
\APACinsertmetastar {%
wang_superglue_2019}%
\begin{APACrefauthors}%
Wang, A.%
, Pruksachatkun, Y.%
, Nangia, N.%
, Singh, A.%
, Michael, J.%
, Hill, F.%
\BDBL {}Bowman, S.%
\end{APACrefauthors}%
\unskip\
\newblock
\APACrefYearMonthDay{2019}{}{}.
\newblock
{\BBOQ}\APACrefatitle {{SuperGLUE}: {A} {Stickier} {Benchmark} for
  {General}-{Purpose} {Language} {Understanding} {Systems}} {{SuperGLUE}: {A}
  {Stickier} {Benchmark} for {General}-{Purpose} {Language} {Understanding}
  {Systems}}.{\BBCQ}
\newblock
\BIn{} H.~Wallach, H.~Larochelle, A.~Beygelzimer, F\BPBI d.~Alché-Buc,
  E.~Fox\BCBL {}\ \BBA {} R.~Garnett\ (\BEDS), \APACrefbtitle {Advances in
  {Neural} {Information} {Processing} {Systems}} {Advances in {Neural}
  {Information} {Processing} {Systems}}\ (\BVOL~32).
\newblock
\APACaddressPublisher{}{Curran Associates, Inc.}
\newblock
\begin{APACrefURL}
  \url{https://proceedings.neurips.cc/paper_files/paper/2019/file/4496bf24afe7fab6f046bf4923da8de6-Paper.pdf}
  \end{APACrefURL}
\PrintBackRefs{\CurrentBib}

\bibitem [\protect \citeauthoryear {%
Y.~Wang%
\ \protect \BOthers {.}}{%
Y.~Wang%
\ \protect \BOthers {.}}{%
{\protect \APACyear {2023}}%
}]{%
wang_self-instruct_2023}
\APACinsertmetastar {%
wang_self-instruct_2023}%
\begin{APACrefauthors}%
Wang, Y.%
, Kordi, Y.%
, Mishra, S.%
, Liu, A.%
, Smith, N\BPBI A.%
, Khashabi, D.%
\BCBL {}\ \BBA {} Hajishirzi, H.%
\end{APACrefauthors}%
\unskip\
\newblock
\APACrefYearMonthDay{2023}{{\APACmonth{07}}}{}.
\newblock
{\BBOQ}\APACrefatitle {Self-{Instruct}: {Aligning} {Language} {Models} with
  {Self}-{Generated} {Instructions}} {Self-{Instruct}: {Aligning} {Language}
  {Models} with {Self}-{Generated} {Instructions}}.{\BBCQ}
\newblock
\BIn{} \APACrefbtitle {Proceedings of the 61st {Annual} {Meeting} of the
  {Association} for {Computational} {Linguistics} ({Volume} 1: {Long}
  {Papers})} {Proceedings of the 61st {Annual} {Meeting} of the {Association}
  for {Computational} {Linguistics} ({Volume} 1: {Long} {Papers})}\ (\BPGS\
  13484--13508).
\newblock
\APACaddressPublisher{Toronto, Canada}{Association for Computational
  Linguistics}.
\newblock
\begin{APACrefURL}
  [{2023-07-20}]\url{https://aclanthology.org/2023.acl-long.754}
  \end{APACrefURL}
\PrintBackRefs{\CurrentBib}

\bibitem [\protect \citeauthoryear {%
Wenger%
}{%
Wenger%
}{%
{\protect \APACyear {2023}}%
}]{%
wenger_ai_2023}
\APACinsertmetastar {%
wenger_ai_2023}%
\begin{APACrefauthors}%
Wenger, J.%
\end{APACrefauthors}%
\unskip\
\newblock
\APACrefYearMonthDay{2023}{}{}.
\newblock
\APACrefbtitle {{AI} {Email} {Summaries}: {Read} emails in seconds.} {{AI}
  {Email} {Summaries}: {Read} emails in seconds.}
\newblock
\begin{APACrefURL}
  [{2023-03-01}]\url{https://www.shortwave.com/blog/ai-email-summaries/}
  \end{APACrefURL}
\PrintBackRefs{\CurrentBib}

\bibitem [\protect \citeauthoryear {%
Wu%
\ \protect \BOthers {.}}{%
Wu%
\ \protect \BOthers {.}}{%
{\protect \APACyear {2021}}%
}]{%
wu_recursively_2021}
\APACinsertmetastar {%
wu_recursively_2021}%
\begin{APACrefauthors}%
Wu, J.%
, Ouyang, L.%
, Ziegler, D\BPBI M.%
, Stiennon, N.%
, Lowe, R.%
, Leike, J.%
\BCBL {}\ \BBA {} Christiano, P.%
\end{APACrefauthors}%
\unskip\
\newblock
\APACrefYearMonthDay{2021}{{\APACmonth{09}}}{}.
\newblock
\APACrefbtitle {Recursively {Summarizing} {Books} with {Human} {Feedback}.}
  {Recursively {Summarizing} {Books} with {Human} {Feedback}.}
\newblock
\APACaddressPublisher{}{arXiv}.
\newblock
\begin{APACrefURL} [{2023-03-07}]\url{http://arxiv.org/abs/2109.10862}
  \end{APACrefURL}
\newblock
\APACrefnote{arXiv:2109.10862 [cs]}
\newblock
\begin{APACrefDOI} \doi{10.48550/arXiv.2109.10862} \end{APACrefDOI}
\PrintBackRefs{\CurrentBib}

\bibitem [\protect \citeauthoryear {%
Xiao%
, Beltagy%
, Carenini%
\BCBL {}\ \BBA {} Cohan%
}{%
Xiao%
\ \protect \BOthers {.}}{%
{\protect \APACyear {2022}}%
}]{%
xiao_primera_2022}
\APACinsertmetastar {%
xiao_primera_2022}%
\begin{APACrefauthors}%
Xiao, W.%
, Beltagy, I.%
, Carenini, G.%
\BCBL {}\ \BBA {} Cohan, A.%
\end{APACrefauthors}%
\unskip\
\newblock
\APACrefYearMonthDay{2022}{{\APACmonth{05}}}{}.
\newblock
{\BBOQ}\APACrefatitle {{PRIMERA}: {Pyramid}-based {Masked} {Sentence}
  {Pre}-training for {Multi}-document {Summarization}} {{PRIMERA}:
  {Pyramid}-based {Masked} {Sentence} {Pre}-training for {Multi}-document
  {Summarization}}.{\BBCQ}
\newblock
\BIn{} \APACrefbtitle {Proceedings of the 60th {Annual} {Meeting} of the
  {Association} for {Computational} {Linguistics} ({Volume} 1: {Long}
  {Papers})} {Proceedings of the 60th {Annual} {Meeting} of the {Association}
  for {Computational} {Linguistics} ({Volume} 1: {Long} {Papers})}\ (\BPGS\
  5245--5263).
\newblock
\APACaddressPublisher{Dublin, Ireland}{Association for Computational
  Linguistics}.
\newblock
\begin{APACrefURL}
  [{2023-07-20}]\url{https://aclanthology.org/2022.acl-long.360}
  \end{APACrefURL}
\newblock
\begin{APACrefDOI} \doi{10.18653/v1/2022.acl-long.360} \end{APACrefDOI}
\PrintBackRefs{\CurrentBib}

\bibitem [\protect \citeauthoryear {%
Xiao%
, Xie%
, Carenini%
\BCBL {}\ \BBA {} He%
}{%
Xiao%
\ \protect \BOthers {.}}{%
{\protect \APACyear {2023}}%
}]{%
xiao_chatgpt-steered_2023}
\APACinsertmetastar {%
xiao_chatgpt-steered_2023}%
\begin{APACrefauthors}%
Xiao, W.%
, Xie, Y.%
, Carenini, G.%
\BCBL {}\ \BBA {} He, P.%
\end{APACrefauthors}%
\unskip\
\newblock
\APACrefYearMonthDay{2023}{{\APACmonth{05}}}{}.
\newblock
\APACrefbtitle {{ChatGPT}-steered {Editing} {Instructor} for {Customization} of
  {Abstractive} {Summarization}.} {{ChatGPT}-steered {Editing} {Instructor} for
  {Customization} of {Abstractive} {Summarization}.}
\newblock
\APACaddressPublisher{}{arXiv}.
\newblock
\begin{APACrefURL} [{2023-05-08}]\url{http://arxiv.org/abs/2305.02483}
  \end{APACrefURL}
\newblock
\APACrefnote{arXiv:2305.02483 [cs]}
\PrintBackRefs{\CurrentBib}

\bibitem [\protect \citeauthoryear {%
Yang%
, Li%
, Zhang%
, Chen%
\BCBL {}\ \BBA {} Cheng%
}{%
Yang%
\ \protect \BOthers {.}}{%
{\protect \APACyear {2023}}%
}]{%
yang_exploring_2023}
\APACinsertmetastar {%
yang_exploring_2023}%
\begin{APACrefauthors}%
Yang, X.%
, Li, Y.%
, Zhang, X.%
, Chen, H.%
\BCBL {}\ \BBA {} Cheng, W.%
\end{APACrefauthors}%
\unskip\
\newblock
\APACrefYearMonthDay{2023}{{\APACmonth{02}}}{}.
\newblock
\APACrefbtitle {Exploring the {Limits} of {ChatGPT} for {Query} or
  {Aspect}-based {Text} {Summarization}.} {Exploring the {Limits} of {ChatGPT}
  for {Query} or {Aspect}-based {Text} {Summarization}.}
\newblock
\APACaddressPublisher{}{arXiv}.
\newblock
\begin{APACrefURL} [{2023-02-17}]\url{http://arxiv.org/abs/2302.08081}
  \end{APACrefURL}
\newblock
\APACrefnote{arXiv:2302.08081 [cs]}
\newblock
\begin{APACrefDOI} \doi{10.48550/arXiv.2302.08081} \end{APACrefDOI}
\PrintBackRefs{\CurrentBib}

\bibitem [\protect \citeauthoryear {%
Yuan%
, Neubig%
\BCBL {}\ \BBA {} Liu%
}{%
Yuan%
\ \protect \BOthers {.}}{%
{\protect \APACyear {2021}}%
}]{%
yuan_bartscore_2021}
\APACinsertmetastar {%
yuan_bartscore_2021}%
\begin{APACrefauthors}%
Yuan, W.%
, Neubig, G.%
\BCBL {}\ \BBA {} Liu, P.%
\end{APACrefauthors}%
\unskip\
\newblock
\APACrefYearMonthDay{2021}{}{}.
\newblock
{\BBOQ}\APACrefatitle {{BARTScore}: {Evaluating} {Generated} {Text} as {Text}
  {Generation}} {{BARTScore}: {Evaluating} {Generated} {Text} as {Text}
  {Generation}}.{\BBCQ}
\newblock
\BIn{} M.~Ranzato, A.~Beygelzimer, Y.~Dauphin, P\BPBI S.~Liang\BCBL {}\ \BBA {}
  J\BPBI W.~Vaughan\ (\BEDS), \APACrefbtitle {Advances in {Neural}
  {Information} {Processing} {Systems}} {Advances in {Neural} {Information}
  {Processing} {Systems}}\ (\BVOL~34, \BPGS\ 27263--27277).
\newblock
\APACaddressPublisher{}{Curran Associates, Inc.}
\newblock
\begin{APACrefURL}
  \url{https://proceedings.neurips.cc/paper_files/paper/2021/file/e4d2b6e6fdeca3e60e0f1a62fee3d9dd-Paper.pdf}
  \end{APACrefURL}
\PrintBackRefs{\CurrentBib}

\bibitem [\protect \citeauthoryear {%
D.~Zhang%
\ \protect \BOthers {.}}{%
D.~Zhang%
\ \protect \BOthers {.}}{%
{\protect \APACyear {2022}}%
}]{%
zhang_ai_2022}
\APACinsertmetastar {%
zhang_ai_2022}%
\begin{APACrefauthors}%
Zhang, D.%
, Maslej, N.%
, Brynjolfsson, E.%
, Etchemendy, J.%
, Lyons, T.%
, Manyika, J.%
\BDBL {}Perrault, R.%
\end{APACrefauthors}%
\unskip\
\newblock
\APACrefYearMonthDay{2022}{{\APACmonth{05}}}{}.
\newblock
\APACrefbtitle {The {AI} {Index} 2022 {Annual} {Report}.} {The {AI} {Index}
  2022 {Annual} {Report}.}
\newblock
\APACaddressPublisher{}{arXiv}.
\newblock
\begin{APACrefURL} [{2023-02-28}]\url{http://arxiv.org/abs/2205.03468}
  \end{APACrefURL}
\newblock
\APACrefnote{arXiv:2205.03468 [cs]}
\newblock
\begin{APACrefDOI} \doi{10.48550/arXiv.2205.03468} \end{APACrefDOI}
\PrintBackRefs{\CurrentBib}

\bibitem [\protect \citeauthoryear {%
J.~Zhang%
, Zhao%
, Saleh%
\BCBL {}\ \BBA {} Liu%
}{%
J.~Zhang%
\ \protect \BOthers {.}}{%
{\protect \APACyear {2020}}%
}]{%
zhang_pegasus_2020}
\APACinsertmetastar {%
zhang_pegasus_2020}%
\begin{APACrefauthors}%
Zhang, J.%
, Zhao, Y.%
, Saleh, M.%
\BCBL {}\ \BBA {} Liu, P.%
\end{APACrefauthors}%
\unskip\
\newblock
\APACrefYearMonthDay{2020}{{\APACmonth{11}}}{}.
\newblock
{\BBOQ}\APACrefatitle {{PEGASUS}: {Pre}-training with {Extracted}
  {Gap}-sentences for {Abstractive} {Summarization}} {{PEGASUS}: {Pre}-training
  with {Extracted} {Gap}-sentences for {Abstractive} {Summarization}}.{\BBCQ}
\newblock
\BIn{} \APACrefbtitle {Proceedings of the 37th {International} {Conference} on
  {Machine} {Learning}} {Proceedings of the 37th {International} {Conference}
  on {Machine} {Learning}}\ (\BPGS\ 11328--11339).
\newblock
\APACaddressPublisher{}{PMLR}.
\newblock
\begin{APACrefURL}
  [{2023-07-20}]\url{https://proceedings.mlr.press/v119/zhang20ae.html}
  \end{APACrefURL}
\newblock
\APACrefnote{ISSN: 2640-3498}
\PrintBackRefs{\CurrentBib}

\bibitem [\protect \citeauthoryear {%
S.~Zhang%
\ \protect \BOthers {.}}{%
S.~Zhang%
\ \protect \BOthers {.}}{%
{\protect \APACyear {2022}}%
}]{%
zhang_opt_2022}
\APACinsertmetastar {%
zhang_opt_2022}%
\begin{APACrefauthors}%
Zhang, S.%
, Roller, S.%
, Goyal, N.%
, Artetxe, M.%
, Chen, M.%
, Chen, S.%
\BDBL {}Zettlemoyer, L.%
\end{APACrefauthors}%
\unskip\
\newblock
\APACrefYearMonthDay{2022}{{\APACmonth{06}}}{}.
\newblock
\APACrefbtitle {{OPT}: {Open} {Pre}-trained {Transformer} {Language} {Models}.}
  {{OPT}: {Open} {Pre}-trained {Transformer} {Language} {Models}.}
\newblock
\APACaddressPublisher{}{arXiv}.
\newblock
\begin{APACrefURL} [{2023-02-28}]\url{http://arxiv.org/abs/2205.01068}
  \end{APACrefURL}
\newblock
\APACrefnote{arXiv:2205.01068 [cs]}
\newblock
\begin{APACrefDOI} \doi{10.48550/arXiv.2205.01068} \end{APACrefDOI}
\PrintBackRefs{\CurrentBib}

\bibitem [\protect \citeauthoryear {%
T.~Zhang%
, Kishore%
, Wu%
, Weinberger%
\BCBL {}\ \BBA {} Artzi%
}{%
T.~Zhang%
\ \protect \BOthers {.}}{%
{\protect \APACyear {2019}}%
}]{%
zhang_bertscore_2019}
\APACinsertmetastar {%
zhang_bertscore_2019}%
\begin{APACrefauthors}%
Zhang, T.%
, Kishore, V.%
, Wu, F.%
, Weinberger, K\BPBI Q.%
\BCBL {}\ \BBA {} Artzi, Y.%
\end{APACrefauthors}%
\unskip\
\newblock
\APACrefYearMonthDay{2019}{{\APACmonth{09}}}{}.
\newblock
{\BBOQ}\APACrefatitle {{BERTScore}: {Evaluating} {Text} {Generation} with
  {BERT}} {{BERTScore}: {Evaluating} {Text} {Generation} with {BERT}}.{\BBCQ}.
\newblock
\begin{APACrefURL}
  [{2023-07-27}]\url{https://openreview.net/forum?id=SkeHuCVFDr}
  \end{APACrefURL}
\PrintBackRefs{\CurrentBib}

\bibitem [\protect \citeauthoryear {%
T.~Zhang%
\ \protect \BOthers {.}}{%
T.~Zhang%
\ \protect \BOthers {.}}{%
{\protect \APACyear {2023}}%
}]{%
zhang_benchmarking_2023}
\APACinsertmetastar {%
zhang_benchmarking_2023}%
\begin{APACrefauthors}%
Zhang, T.%
, Ladhak, F.%
, Durmus, E.%
, Liang, P.%
, McKeown, K.%
\BCBL {}\ \BBA {} Hashimoto, T\BPBI B.%
\end{APACrefauthors}%
\unskip\
\newblock
\APACrefYearMonthDay{2023}{{\APACmonth{01}}}{}.
\newblock
\APACrefbtitle {Benchmarking {Large} {Language} {Models} for {News}
  {Summarization}.} {Benchmarking {Large} {Language} {Models} for {News}
  {Summarization}.}
\newblock
\APACaddressPublisher{}{arXiv}.
\newblock
\begin{APACrefURL} [{2023-02-17}]\url{http://arxiv.org/abs/2301.13848}
  \end{APACrefURL}
\newblock
\APACrefnote{arXiv:2301.13848 [cs] version: 1}
\PrintBackRefs{\CurrentBib}

\bibitem [\protect \citeauthoryear {%
Y.~Zhang%
\ \protect \BOthers {.}}{%
Y.~Zhang%
\ \protect \BOthers {.}}{%
{\protect \APACyear {2022}}%
}]{%
zhang_summn_2022}
\APACinsertmetastar {%
zhang_summn_2022}%
\begin{APACrefauthors}%
Zhang, Y.%
, Ni, A.%
, Mao, Z.%
, Wu, C\BPBI H.%
, Zhu, C.%
, Deb, B.%
\BDBL {}Zhang, R.%
\end{APACrefauthors}%
\unskip\
\newblock
\APACrefYearMonthDay{2022}{{\APACmonth{05}}}{}.
\newblock
{\BBOQ}\APACrefatitle {Summ{\textasciicircum}{N}: {A} {Multi}-{Stage}
  {Summarization} {Framework} for {Long} {Input} {Dialogues} and {Documents}}
  {Summ{\textasciicircum}{N}: {A} {Multi}-{Stage} {Summarization} {Framework}
  for {Long} {Input} {Dialogues} and {Documents}}.{\BBCQ}
\newblock
\BIn{} \APACrefbtitle {Proceedings of the 60th {Annual} {Meeting} of the
  {Association} for {Computational} {Linguistics} ({Volume} 1: {Long}
  {Papers})} {Proceedings of the 60th {Annual} {Meeting} of the {Association}
  for {Computational} {Linguistics} ({Volume} 1: {Long} {Papers})}\ (\BPGS\
  1592--1604).
\newblock
\APACaddressPublisher{Dublin, Ireland}{Association for Computational
  Linguistics}.
\newblock
\begin{APACrefURL}
  [{2023-07-20}]\url{https://aclanthology.org/2022.acl-long.112}
  \end{APACrefURL}
\newblock
\begin{APACrefDOI} \doi{10.18653/v1/2022.acl-long.112} \end{APACrefDOI}
\PrintBackRefs{\CurrentBib}

\bibitem [\protect \citeauthoryear {%
I.~Zhao%
}{%
I.~Zhao%
}{%
{\protect \APACyear {2023}}%
}]{%
zhao_notion_2023}
\APACinsertmetastar {%
zhao_notion_2023}%
\begin{APACrefauthors}%
Zhao, I.%
\end{APACrefauthors}%
\unskip\
\newblock
\APACrefYearMonthDay{2023}{{\APACmonth{02}}}{}.
\newblock
\APACrefbtitle {Notion {AI} is here, for everyone.} {Notion {AI} is here, for
  everyone.}
\newblock
\begin{APACrefURL}
  [{2023-03-01}]\url{https://www.notion.so/blog/notion-ai-is-here-for-everyone}
  \end{APACrefURL}
\PrintBackRefs{\CurrentBib}

\bibitem [\protect \citeauthoryear {%
Y.~Zhao%
, Saleh%
\BCBL {}\ \BBA {} Liu%
}{%
Y.~Zhao%
\ \protect \BOthers {.}}{%
{\protect \APACyear {2020}}%
}]{%
zhao_seal_2020}
\APACinsertmetastar {%
zhao_seal_2020}%
\begin{APACrefauthors}%
Zhao, Y.%
, Saleh, M.%
\BCBL {}\ \BBA {} Liu, P\BPBI J.%
\end{APACrefauthors}%
\unskip\
\newblock
\APACrefYearMonthDay{2020}{{\APACmonth{06}}}{}.
\newblock
\APACrefbtitle {{SEAL}: {Segment}-wise {Extractive}-{Abstractive} {Long}-form
  {Text} {Summarization}.} {{SEAL}: {Segment}-wise {Extractive}-{Abstractive}
  {Long}-form {Text} {Summarization}.}
\newblock
\APACaddressPublisher{}{arXiv}.
\newblock
\begin{APACrefURL} [{2023-03-20}]\url{http://arxiv.org/abs/2006.10213}
  \end{APACrefURL}
\newblock
\APACrefnote{arXiv:2006.10213 [cs]}
\PrintBackRefs{\CurrentBib}

\bibitem [\protect \citeauthoryear {%
Zhong%
, Liu%
, Wang%
, Qiu%
\BCBL {}\ \BBA {} Huang%
}{%
Zhong%
\ \protect \BOthers {.}}{%
{\protect \APACyear {2019}}%
}]{%
zhong_searching_2019}
\APACinsertmetastar {%
zhong_searching_2019}%
\begin{APACrefauthors}%
Zhong, M.%
, Liu, P.%
, Wang, D.%
, Qiu, X.%
\BCBL {}\ \BBA {} Huang, X.%
\end{APACrefauthors}%
\unskip\
\newblock
\APACrefYearMonthDay{2019}{{\APACmonth{07}}}{}.
\newblock
{\BBOQ}\APACrefatitle {Searching for {Effective} {Neural} {Extractive}
  {Summarization}: {What} {Works} and {What}'s {Next}} {Searching for
  {Effective} {Neural} {Extractive} {Summarization}: {What} {Works} and
  {What}'s {Next}}.{\BBCQ}
\newblock
\BIn{} \APACrefbtitle {Proceedings of the 57th {Annual} {Meeting} of the
  {Association} for {Computational} {Linguistics}} {Proceedings of the 57th
  {Annual} {Meeting} of the {Association} for {Computational} {Linguistics}}\
  (\BPGS\ 1049--1058).
\newblock
\APACaddressPublisher{Florence, Italy}{Association for Computational
  Linguistics}.
\newblock
\begin{APACrefURL} [{2023-07-20}]\url{https://aclanthology.org/P19-1100}
  \end{APACrefURL}
\newblock
\begin{APACrefDOI} \doi{10.18653/v1/P19-1100} \end{APACrefDOI}
\PrintBackRefs{\CurrentBib}

\end{thebibliography}

\end{document}